\documentclass[10pt,journal]{IEEEtran}
\usepackage{amsmath,amsfonts}
\usepackage{array}
\usepackage{textcomp}
\usepackage{stfloats}
\usepackage{url}
\usepackage{verbatim}
\usepackage{graphicx}
\usepackage{cite}

\usepackage{amsmath,amssymb,amsfonts}%
\usepackage{amsthm}%
\usepackage{mathrsfs}%
\usepackage{adjustbox}
\usepackage{xcolor}%
\usepackage{textcomp}%
\usepackage{manyfoot}%
\usepackage{booktabs}%
\usepackage{algorithmicx}%
\usepackage{algpseudocode}%
\usepackage{listings}%

\usepackage{amsmath,amssymb,amsfonts}
\usepackage{color}
\usepackage{makecell,diagbox}
\usepackage{multirow}
\usepackage{booktabs}
\usepackage{float} 
\usepackage{subfigure}
\usepackage{tabularx}
\usepackage{pdfpages}
\usepackage[linesnumbered,ruled,vlined,boxed]{algorithm2e}
\usepackage{graphics}
\usepackage{epsfig}
\usepackage{lineno,hyperref}
\usepackage{dsfont}
\begin{document}
\title{Vertical Federated Unlearning via Backdoor Certification}

\author{Mengde Han, Tianqing Zhu*, Lefeng Zhang, Huan Huo, Wanlei Zhou

\thanks{Mengde Han and Huan Huo are with the School of Computer Science, University of Technology Sydney,
Ultimo, 2007 NSW, Australia. Email: Mengde.Han@student.uts.edu.au, Huan.Huo@uts.edu.au}
\thanks{Tianqing Zhu, Lefeng Zhang and Wanlei Zhou are with the Faculty of Data Science, City University of Macau,
Macau, China. Email: tqzhu@cityu.edu.mo, lfzhang@cityu.edu.mo, wlzhou@cityu.edu.mo}
\thanks{This paper is supported by the Australian Research Council Discovery DP200100946 and DP230100246.}
\thanks{(Corresponding Author: Tianqing Zhu)}
}
\markboth{Journal of \LaTeX\ Class Files}
{Shell \MakeLowercase{\textit{et al.}}: A Sample Article Using IEEEtran.cls for IEEE Journals}

\IEEEpubid{0000--0000/00\$00.00~\copyright~2024 IEEE}

\maketitle

\begin{abstract}
Vertical Federated Learning (VFL) offers a novel paradigm in machine learning, enabling distinct entities to train models cooperatively while maintaining data privacy. This method is particularly pertinent when entities possess datasets with identical sample identifiers but diverse attributes. Recent privacy regulations emphasize an individual's \emph{right to be forgotten}, which necessitates the ability for models to unlearn specific training data. The primary challenge is to develop a mechanism to eliminate the influence of a specific client from a model without erasing all relevant data from other clients. Our research investigates the removal of a single client's contribution within the VFL framework. We introduce an innovative modification to traditional VFL by employing a mechanism that inverts the typical learning trajectory with the objective of extracting specific data contributions. This approach seeks to optimize model performance using gradient ascent, guided by a pre-defined constrained model. We also introduce a backdoor mechanism to verify the effectiveness of the unlearning procedure. Our method avoids fully accessing the initial training data and avoids storing parameter updates. Empirical evidence shows that the results align closely with those achieved by retraining from scratch. Utilizing gradient ascent, our unlearning approach addresses key challenges in VFL, laying the groundwork for future advancements in this domain. All the code and implementations related to this paper are publicly available at https://github.com/mengde-han/VFL-unlearn.
\end{abstract}

\begin{IEEEkeywords}
Federated Learning, Vertical Federated Learning, Federated Unlearning, Gradient Ascent
\end{IEEEkeywords}

    \section{Introduction}
    \IEEEPARstart{I}n the dynamic realm of machine learning, Federated Learning (FL) stands out as a key approach to address the challenges of data privacy, emphasizing data decentralization at the source while promoting collaborative model development. Conventional FL predominantly functions in a horizontal framework, where different organizations have data samples from identical feature spaces but vary in their distributions. However, with diverse data origins and the intricacies of real-world datasets, it becomes evident that many fields feature vertically partitioned data. To cater to this distinct requirement, vertical federated learning (VFL) was devised. VFL encourages joint model training among parties with different attributes of shared entities, broadening the feature domain without undermining data confidentiality.
\IEEEpubidadjcol    
The recent legal requirements, such as the \emph{right to be forgotten}, motivated a new requirement named machine unlearning, which requires the unlearned model to behave as if it has never processed the specified data. In VFL, merely unlearning at a local level might not be comprehensive as other federated participants could still have remnants of the removed data. In VFL setups, where participants share the same user domain but contribute unique feature sets, unlearning a client's contribution entails removing that client’s features from the overall feature space. This process contrasts with Horizontal Federated Learning (HFL), where unlearning focuses on completely excluding specific data samples from the model. Consequently, existing unlearning methods in HFL, such as FedEraser~\cite{liu2020federated} and Forget-SVGD\cite{gong2022forget}, are not directly applicable to VFL scenarios. The misalignment stems from the fundamental difference in the unlearning focus: in VFL, the emphasis is on erasing specific feature contributions tied to user data, rather than simply removing entire data samples. This distinction introduces unique challenges in VFL, as eliminating features must also ensure that the learned relationships and model behavior associated with those features are effectively forgotten. Therefore, specialized unlearning algorithms tailored for VFL are essential to address these challenges comprehensively. Developing such methods remains a crucial area for further research, as effective unlearning in VFL is critical for enhancing both data privacy and model integrity.
Specifically, there are two challenges associated with unlearning a client's data:
\begin{itemize}
    \item How can we efficiently diminish the impact of the target client's data on a selected party, rather than eliminating the client's entire contribution?
    \item How can we effectively verify that the proposed unlearning mechanism really unlearns the influence of the targeted data?
\end{itemize}

To address these challenges, in this paper, we outlined a procedure to eliminate a client's influence from a trained model, ensuring 'unlearning' without compromising the integrity of the existing model. Specifically, we developed a novel unlearning framework for VFL that consists of the unlearning process and verification process. For the unlearning process, we employed the gradient ascent method alongside a constrained model, which is derived from the vertical federated learning training procedure. The unlearning procedure is designed to reverse the training process, enabling the removal of the target client's information from the selected party without requiring the storage of intermediate gradients. To verify the unlearning, we introduced a backdoor-based method, which injects backdoors into the dataset of the target client associated with the selected party. The decreasing in backdoor accuracy indicates a successful unlearning of the targeted data. Moreover, we integrated membership inference attacks (MIA)~\cite{yeom2018privacy} as an additional component for verification. If the MIA demonstrates that the model can no longer accurately predict the membership of the target client's data, it further verifies the efficacy of our unlearning process.


Our paper presents the following contributions: First of all, it contains most of shadow model that

\begin{itemize}

\item We propose a novel method for vertical federated unlearning. This technique employs constrained gradient ascent on the targeted client, counteracting the effects of their data on a selected party. Following the exclusion of the target client's data, the model demonstrates efficient convergence, signifying the emergence of an effective unlearning algorithm in VFL on image datasets.

\item We develop a gradient ascent based metric to evaluate the efficacy of the unlearning process in the context of VFL. The metric is crucial in scenarios where a target client introduces malicious data during the training phase with a selected party. By using this metric, we can assess the capability of the proposed method in counteracting the negative effects of such data intrusions. 
This evaluation method shows that our unlearning procedure effectively removes the target client’s data and, importantly, nullifies any detrimental effects stemming from data-poisoning attacks from that client.

\item We conduct thorough empirical simulations using three real-world datasets and two different metrics to evaluate the efficacy of our suggested method. The experimental results demonstrate that our approach not only effectively removes influence of the target client's data from the selected party but also restores the global model's accuracy with a few post-training iterations. Furthermore, the performance of our unlearning method is comparable to that achieved by retraining the model from scratch. This outcome highlights significant efficiency, especially when compared to the expensive alternatives of retraining-based solutions.

\end{itemize}

	\section{Related Works} \label{Related}

    \subsection{Vertical Federated Learning}
Vertical Federated Learning (VFL) is especially apt for FL scenarios where participants have datasets with matching sample IDs but distinct attribute spaces. The advent of VFL can be traced back to the seminal work by Hardy et al.\cite{hardy2017private}. They proposed a federated logistic regression model using encrypted messages via an additively homomorphic scheme, and further explored the effects of entity resolution errors on learning results.

Taking a unique direction, Yang et al.\cite{yang2019federated} designed a parallel distributed system that forgoes the necessity of a third-party coordinator. Their design, which aims to decrease data leakage threats and simplify the system, is grounded on the notion of pre-aligned sample IDs. Probing further into VFL nuances, Wang et al.\cite{wang2019measure} developed a means to determine feature significance in local datasets owned by VFL participants. By leveraging a Shapley Value-focused strategy, their method dynamically omits various feature sets and gauges the resulting effects on the FL model's performance.

In terms of privacy conservation, Wu et al.\cite{wu2020privacy} introduced Pivot, a state-of-the-art solution tailored for private vertical decision tree training and prediction. Pivot guarantees that only collaboratively agreed-upon intermediate data is exchanged between clients. 
Adding to this innovation, Jiang et al.\cite{jiang2022vf} unveiled VF-MINE, a fresh mutual information estimator based on Fagin’s algorithm. Building on this, they crafted a group testing-oriented framework, VF-PS, which marries homomorphic encryption with batching optimization. Addressing privacy concerns, there are also comprehensive surveys on differential privacy, including various attack and defense mechanisms, as discussed in recent studies\cite{zhou2022adversarial}\cite{zhu2020more}. Addressing fairness in VFL, Qi et al.\cite{qi2022fairvfl} formulated a model known as FairVFL. This construct enhances VFL fairness by promoting uniform and impartial sample depictions, all while maintaining data confidentiality. Pushing the envelope further in feature selection and privacy, Li et al.\cite{li2023fedsdg} introduced the Federated Stochastic Dual-Gate based Feature Selection (FedSDGFS) method. This technique uses a Gaussian stochastic dual-gate for adept approximation of feature selection probabilities. Notably, their methodology assures privacy through Partially Homomorphic Encryption, obviating the need for a trusted intermediary.

    \subsection{Machine Unlearning}
    Machine unlearning, an emerging and vital field within artificial intelligence, has been the focus of numerous pivotal studies. These investigations delve into unlearning challenges across various machine learning scenarios. The genesis of machine unlearning traces back to Cao and Yang\cite{cao2015towards}. They pioneered an algorithm that adeptly adjusted the optimization objective function, providing a swift alternative to retraining models from the ground up by facilitating efficient data sample removal. Building on this foundation, Du et al.\cite{du2019lifelong} proposed solutions to combat increasing loss and catastrophic forgetting. Employing generative models, their design could morph a range of deep learning-based anomaly detection algorithms into their lifelong versions.

The field of clustering saw advancements in unlearning as well. Notably, Ginart et al.\cite{ginart2019making} developed dual unlearning techniques for the $k$-means clustering algorithm, harnessing model compression methods to effectively delete specific data samples from clusters. Bourtoule et al.\cite{bourtoule2021machine} unveiled the SISA training framework, emphasizing reducing computational overhead even at the cost of additional space. By ingeniously segmenting datasets, it facilitated selective retraining post the unlearning process. Neel et al.\cite{neel2021descent} made groundbreaking contributions by crafting data deletion algorithms capable of withstanding numerous adversarial updates. Integrating convex optimization with reservoir sampling, their methods ensured consistent runtime and maintained steady-state error benchmarks.

Sekhari et al.\cite{sekhari2021remember} introduced an unlearning method to pinpoint the best solutions for strongly convex challenges. By incorporating an uncertainty factor into the optimizer, their approach achieved nuanced data excision from models. Thudi et al.\cite{thudi2022necessity} took a philosophical angle, asserting that unlearning is purely algorithmic. They argued that genuine unlearning can only be evidenced through the execution of a distinct algorithm crafted explicitly for the task.

Recently, Tarun et al.\cite{tarun2023fast} unveiled a transformative machine unlearning paradigm, marked by error-maximizing noise generation and an innovative weight-altering scheme, adeptly addressing dominant issues. Simultaneously, Chundawat et al.\cite{chundawat2023zero} brought forward a groundbreaking perspective on machine unlearning, focusing on data-absent scenarios, thus catering to extreme, yet plausible, use cases. Recently, Xu et al.\cite{xu2023machine} conducted an analysis and comparison of the feasibility of current unlearning solutions across various scenarios. In an effort to untangle knowledge linkages, Lin et al.\cite{lin2023erm} introduced the entanglement-reduced mask (ERM). Along with the specially crafted masks in ERM-KTP, this structure enhances interpretability, revealing both the unlearning processes and their effects on specific data points.

Despite these substantial contributions, a void persists in tailoring these algorithms for vertical federated unlearning. While most existing methods are preoccupied with eradicating whole data samples, the nuanced challenge of removing particular user data features within models demands a bespoke strategy in vertical federated contexts. 
    
    \subsection{Federated Unlearning}
 In the domain of federated learning, there have been several significant contributions. Gong et al.\cite{gong2022forget} introduced the Forget-SVGD algorithm, a specialized federated unlearning technique tailored for Bayesian models. This approach enhances the efficiency of the unlearning process by employing the Stein Variational Gradient Descent (SVGD) method. Addressing the challenge of unlearning in FL, Liu et al.\cite{liu2020federated} adopted the gradient ascent method. Their approach trained the model backward but had the downside of increased communication overhead. On a related note, Wu et al.\cite{wu2022federated} significantly accelerated the training process in federated unlearning by incorporating knowledge distillation techniques, especially following the removal of data from target clients. In a parallel development, Wei et al.\cite{wei2021user} devised a sophisticated user-level privacy protection scheme, utilizing the principles of differential privacy technology to enhance data security.

Yuan et al.\cite{yuan2023federated} revealed FRU, a method designed for unlearning in federated recommendation systems. They drew inspiration from the rollback mechanism seen in database management systems based on transaction logs. Taking a different angle, Zhu et al.\cite{zhu2023heterogeneous} introduced FedLU. This innovative method, designed for heterogeneous knowledge graph embedding, integrates an unlearning technique rooted in cognitive neuroscience. Su et al.\cite{su2023asynchronous} developed Knot, a unique clustered aggregation mechanism. It's tailored for asynchronous federated learning and operates on the premise that clients can be clustered, with aggregation limited to each cluster. This confines the need for retraining after data erasure. Zhang et al.\cite{zhang2023fedrecovery} proposed FedRecovery, a method anchored in differential privacy. It counteracts a client's influence by removing a weighted sum of gradient residuals from the global model. Additionally, it employs customized Gaussian noise, ensuring the unlearned and retrained models are statistically indistinguishable. 

While much of the current research is centered on Horizontal Federated Learning (HFL), Deng et al.\cite{deng2023vertical} distinguish themselves. They've proposed a method specifically for vertical federated unlearning. This technique imposes constraints on intermediate parameters and deducts updates from the target client from the global model. However, it's essential to note that this approach is strictly for tabular data. This highlights our proposal as the trailblazing vertical federated unlearning technique for image data.

\subsection{Verification of Machine Unlearning}

For the verification of machine unlearning, Gu et al.\cite{gu2017badnets} spotlighted potential security vulnerabilities in outsourced training. In such a scenario, an adversary might construct a 'BadNet' or a backdoored neural network. While this network exhibits superior performance on typical user samples, it can malfunction when presented with certain inputs. In a related vein, Fang et al.\cite{fang2020local} proposed a technique to sabotage the local model in FL by reversing the direction of local model parameter updates. Li et al.\cite{li2020invisible} delved into stealthy attack trigger patterns, emphasizing their elusiveness to human detection, thereby enhancing the efficacy of backdoor attacks. Tolpegin et al.\cite{tolpegin2020data}, Wan et al.\cite{wan2024data}, and Zhang et al.\cite{zhang2021deep} disclosed that data-poisoning attacks can significantly impact crucial metrics such as classification accuracy and recall, this being the case even when only a small number of malicious entities are involved. Concurrently, Zhou et al.\cite{zhou2021deep}, and Feng et al.\cite{feng2022detecting} showcased an inventive optimization-driven model poisoning attack, underlining its efficiency, persistence, and stealth.

From a defensive perspective, Sun et al.\cite{sun2021fl} presented FL-WBC, a client-oriented defense mechanism proficient in mitigating the adverse effects of model poisoning attacks on the global model. The essence of FL-WBC lies in identifying and adjusting the parameter space associated with prolonged attack influences during local training. On a similar note, Zhang et al.\cite{zhang2022fldetector} introduced FLDetector, a tool crafted to detect rogue clients. It predicts a client's model update based on past data and subsequently flags anomalies across iterations. There are also several approaches focused on enhancing privacy protection~\cite{zhang2023privacyeafl}\cite{shen2021privacy}. Further deepening this narrative, Chen et al.\cite{chen2023dark} unveiled H-CARS, an avant-garde approach to poisoning recommender systems. This method can invent user profiles and interaction logs by retracing the learning path.

Although extensive research has been conducted in Horizontal Federated Learning (HFL), our study stands apart. We distinctively tailor and adapt attack methodologies for the VFL setting, aiming to assess the resilience of the unlearned model against such attacks.

	\section{Background and Preliminary} \label{Background}

 \subsection{Notations}
    \begin{table}[ht]
    \caption{Notations}
    \label{Table.notation}
    \begin{adjustbox}{width=250pt}
    \centering
    \begin{tabular}{@{}llll@{}}
    \hline
    Symbol & Definition \\
    \hline
    $N$ & the number of clients \\
    $D_i$ & the dataset of client $i$ \\
    $\mathbf{W}$ & the gradient of the model \\
    $\mathbf{W}_A,\mathbf{W}_B,\mathbf{W}_C$ & the gradient of party $A$, party $B$ and coordinator $C$ \\
    $L$ & the objective loss function of the model \\
    $L_A,L_B,L_C$ & the loss function of party $A$, party $B$ and coordinator $C$ \\
    $\oplus$  & concatenation \\
    $E, E_u$  & the global epochs of vertical federated learning and unlearning\\
    $m, m_u$  & the batch size of vertical federated learning and unlearning\\
    $\eta, \eta_u$  & the learning rate of vertical federated learning and unlearning \\
    \hline
    \end{tabular}
    \end{adjustbox}
    \end{table}
    \subsection{Federated Learning}
  Consider a federated learning framework with $N$ clients, each holding a dataset $D_i = \left\{x_i,y_i\right\}$, where $x_i$ denotes the data and $y_i$ represents the corresponding label. The Federated Averaging algorithm (FedAvg)\cite{mcmahan2017communication} is a widely-adopted method in federated learning. The FedAvg learning task involves minimizing the objective loss function $L(\mathbf{W}; x_i, y_i)$, where $\mathbf{W}$ represents the model parameters (gradients). The objective is to achieve empirical minimization $\mathbf{W}^\ast = arg\min_\mathbf{W}\frac{1}{N}\sum_{i=1}^{N}L(\mathbf{W};x_i,y_i)$ over all $N$ clients' data, with $(x_i, y_i)$ signifying the training sample of client $i$.

In each iteration, $N$ clients update their local models based on the globally distributed parameters. After updating, they transmit their local model adjustments back to the server. The server aggregates these updates, calculating the average gradient, which subsequently updates the global model's parameters: $\mathbf{W}^{(k)}\leftarrow\mathbf{W}^{(k-1)}-\eta\frac{1}{N}\sum_{i=1}^{N}\nabla L(\mathbf{W}^{(k-1)};x_i,y_i)$, where $\eta$ denotes the learning rate. Essentially, the FedAvg algorithm functions by selecting clients, having them run $E$ global epochs of SGD on their local data, and subsequently averaging the resulting local models.

\subsection{Vertical Federated Learning and Unlearning}
\begin{figure}[]
\centering
\includegraphics[scale=0.37]{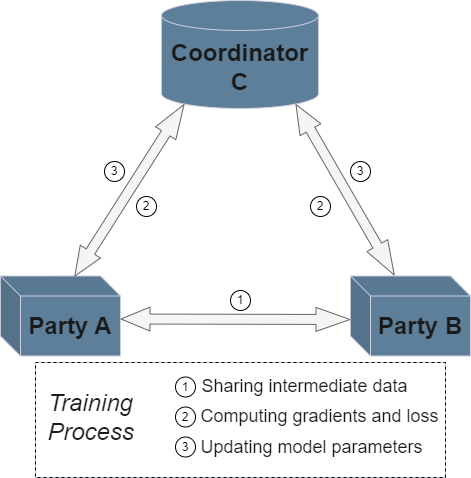}
\caption{The architecture of a vertical federated learning system}
\label{FL overview}
\end{figure}

In vertical federated learning (VFL), the setup diverges considerably from traditional horizontal federated learning. In VFL, each party doesn't have the entire data feature of the client. Rather, every entity in VFL processes distinct feature partitions of a single data sample. Suppose there are two parties, $A$ and $B$. For a data sample $i$, the features $x_i$ are divided into $x_{i}^{A}$ and $x_{i}^{B}$, such that $x_{i} = x_{i}^{A}\oplus x_{i}^{B}$, with 
$\oplus$ symbolizing concatenation. These partitioned features are independently held by $A$ and $B$.

Fig.~\ref{FL overview} depicts a typical VFL setup where $A$ and $B$ collaboratively train a model under the supervision of a coordinator $C$. VFL training has parties exchange intermediate computations to evaluate training losses and gradients while preserving data privacy. The raw data remains localized.

Broadly, model training in VFL consists of three steps: Step 1: $A$ and $B$ encrypt and swap intermediate representations for gradient and loss computations. Step 2: Both $A$ and $B$ determine the encrypted gradients and calculate the loss. They then forward the encrypted gradients and losses to $C$. Step 3: $C$ decrypts the gradients, updates the model parameters, $\mathbf{W}_A$ and $\mathbf{W}_B$, and subsequently sends them back to $A$ and $B$.

In a vertical federated learning training round, each participating client's objective is to develop a local model that effectively minimizes its empirical risk. This can be encapsulated as:
\begin{equation}
    \min _{\mathbf{W}_C \in \mathbb{R}^d} L(\mathbf{W}_C):=\frac{1}{N} \sum_{i=1}^{N} L_{C}\left(\mathbf{W}_{C};x_i, y_i\right)
    \label{lossfunc}
\end{equation}
Here, $L_{C}$ denotes the loss on the coordinator's side for the training sample $(x_i, y_i)$on client $i$ using the model parameters $\mathbf{W}_C$. Further, $L_{C}$is derived by concatenating $L_{A}$ and $L_{B}$, which denote the losses of predictions from parties $A$ and $B$. A more detailed discussion will be presented in the following chapter.

  In federated learning, once the global model is trained collectively by $N$ clients, it embodies the data contributions from all these clients. Federated unlearning emerges as a mechanism that allows the removal or "unlearning" of the contribution of a specific client from this global model. The objective is to produce a model as though it was trained by the remaining $N-1$ clients, ensuring that the model's accuracy and fairness are maintained. This process is vital, especially in situations where there's a need to protect user data, respect user deletion requests, or adhere to data protection regulations.

    The significance of federated unlearning becomes even more prominent in vertical federated learning. In VFL, clients typically have different features for the same set of samples. Each client maintains its dataset privately and runs a localized model, tailored for training collaboratively. In this scenario, unlearning doesn't just mean removing a particular client's data but removing specific features across all shared samples. It's worth noting that strategies used for unlearning in horizontal federated learning, where entire data samples might be removed, don't directly apply to VFL due to this distinction of feature-based sharing.

    \subsection{Verification of Unlearning in VFL}
    
    The process of unlearning in federated learning extends beyond mere data removal; it entails the rigorous evaluation of the extent to which data has been expunged from the model. Various methodologies have been proposed to ensure and measure this unlearning effectively.

Unlearning in a federated learning setup can be conceptually visualized as the process of selectively fine-tuning the range of models generated by the federated learning system. This adjustment is carried out to specifically exclude or minimize the influence of certain data or patterns previously learned during the training phase. Ideally, the difference between models—both with and without the unlearned data—should be minimal when assessed using certain metrics\cite{bourtoule2021machine}. However, the challenge is that evaluating this difference, especially when considering entire distributions, is computationally demanding, rendering it impractical in many situations.

Historically, scholars have circumvented this challenge by exploring alternative assessment criteria. They often juxtapose retrained and unlearned models by inspecting their weights or outputs. Common metrics include the $\ell_2$-distance and the Kullback-Leibler (KL) divergence. There's also a growing emphasis on leveraging privacy-centric metrics, which might measure, for instance, potential privacy breaches or vulnerabilities to specific inference attacks.

A particularly intriguing tool in the federated unlearning toolkit is the use of backdoor triggers\cite{gu2017badnets}. Here, certain data samples are covertly modified, introducing a hidden "backdoor" pattern. The federated model, when trained on this data, becomes subtly but effectively compromised. By analyzing the susceptibility of both the initial and unlearned models to these tampered samples, we can gauge the efficacy of the unlearning process. If an unlearned model proves resistant to these tampered patterns, it suggests that the unlearning was successful, purging the malicious influence while preserving its competence on clean data.

\section{Methodology} \label{Method}
    \subsection{Overview of Vertical Federated Unlearning via Backdoor Certification}
    In this section, we present the overview of our proposed approach. Our objective is to mitigate the impact of a specific target client within a selected party in the context of vertical federated learning. 
    We initially employ the traditional FedAvg\cite{mcmahan2017communication} within vertical federated learning settings. By averaging the weights of all clean clients' models, we obtain the constrained model which then serves as a reference for our unlearning procedure. Subsequently, by using an $\ell_2$-norm ball centered around the constrained model, we apply gradient ascent to the FedAvg model on the target client's selected party to diminish its influence. Given this bounded approach, we can obtain an optimized weight that maximizes the loss of the target client while remaining within a predefined distance from the constrained model. This ensures that the utility of the unlearned model remains robust. After establishing the initial unlearned model, we executed several rounds of global training excluding the target client from the selected party, and the effectiveness of the globally unlearned model notably exceeds that of its locally unlearned counterpart. For the post-trained unlearned model, we contend that its performance is on par with the retrained model. To assess the efficacy of our unlearning method, we incorporate a backdoor mechanism into our evaluation process. To further validate its effectiveness, we also conducted a membership inference attack. We'll delve into the specifics of each step in the subsequent sections.
    
    \subsection{The VFL Framework}
	\begin{figure}[ht]
		\includegraphics[scale=0.45]{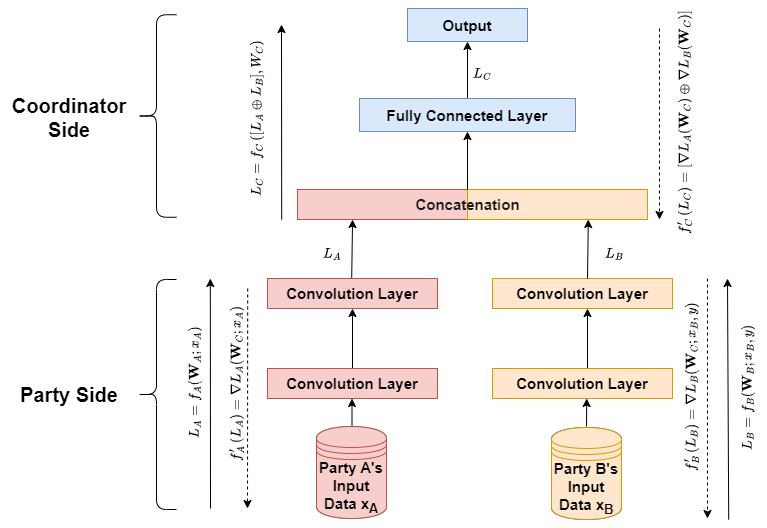}
		\caption{The VFL framework adopted in the paper.}
        \label{SplitNN}
	\end{figure}
    
    In vertical federated learning, various parties function within a shared sample space, yet they handle distinct feature spaces. Each participating entity maintains a local model that is trained using their own private data. This training enables them to compute intermediate parameters at each stage of the learning process. Subsequently, these intermediate parameters are uploaded to a centralized server. The server, upon receiving these parameters, aggregates them and then redistributes the consolidated intermediate parameters back to all the participating parties for further model refinement.
        
    In this paper, we adopts an instance of SplitNN\cite{gupta2018distributed} as the VFL framework, which is illustrated in Figure~\ref{SplitNN}.
    During the forward propagation phase, party $A$ and $B$ calculate the outputs of their respective local models ($\mathbf{W}_A$ and $\mathbf{W}_B$) and transmit them to the coordinator $C$, here we assume label $y$ is held by party $B$:
    \begin{equation}
    L_{A}=f_{A}(\mathbf{W}_A; x_{A})
    \end{equation}
    \begin{equation}
    L_{B}=f_{B}(\mathbf{W}_B; x_{B},y)
    \end{equation}
    The local outputs forwarded, denoted as $L_A$ and $L_B$, are concatenated at the coordinator side and then input into the coordinator model ($\mathbf{W}_C$) to produce a combined inference, resulting in $L_C$.
    \begin{equation}
    L_{C}=f_{C}\left(\left[L_{A} \oplus L_{B}\right], W_{C}\right)
    \end{equation}
    Here, $\oplus$ signifies concatenation. $f$ symbolizes the neural network, and $f^\prime$ denotes its derivative. During the backward propagation phase, the coordinator $C$ calculates the gradients $f^\prime_C(L_C)$, executes gradient descent on its model, and determines the gradients of the local outputs initially forwarded by party $A$ and $B$ as below:
    \begin{equation}
    f_{C}^{\prime}\left(L_{C}\right) = \left[\nabla L_A(\mathbf{W}_{C}) \oplus \nabla L_B(\mathbf{W}_{C})\right]
    \end{equation}
    
   On the party side, upon receiving the gradients, party $A$ and $B$ compute the local gradients $f_{A}^{\prime}\left(L_{A}\right) = \nabla L_A(\mathbf{W}_{C};x_A)$ and $f_{B}^{\prime}\left(L_{B}\right) = \nabla L_B(\mathbf{W}_{C};x_B,y)$, subsequently updating their local models. Notably, throughout the forward and backward propagations, there is no direct transmission of raw data to the coordinator or amongst party $A$ and $B$, and all intermediary data is encrypted. Moreover, by leveraging the SplitNN protocol, more sophisticated models, such as ResNet and LSTM, can be developed following the VFL framework.

    \subsection{Unlearning via Gradient Ascent}
    The primary objective in vertical federated learning is to collaboratively develop a model that effectively minimizes the local empirical risk for every client in party $A$ and $B$, thereby finding an optimized $\mathbf{W}$ that minimizes $L(\mathbf{W})$. 
    Clients perform several passes of stochastic gradient descent locally to discover a model that exhibits low empirical loss. The optimization problem for the training process for client $i$ is defined as follows:
    \begin{equation}
    \min _{\mathbf{W}_{i} \in \mathbb{R}^d} L(\mathbf{W}_{i}):=\frac{1}{\text{Len}(D_i)} \sum_{i\in D_{i}} L\left(\mathbf{W}_{i};x_i, y_i\right)
    \label{lossclient}
    \end{equation}
    where $D_i$ represents the dataset maintained locally by client $i$.
    
    The process of unlearning in VFL requires the careful extraction of a target client's contributions from the shared global model, while keeping intact the inputs from other participating clients. This is particularly challenging due to the limited information shared in VFL, which makes it difficult to clearly differentiate between parameters influenced by the target client and those associated with others. To overcome these challenges, we adopt a novel approach inspired by the work of Halimi et al. \cite{halimi2022federated}, which involves reversing the typical training process. In this reversed approach, during the unlearning phase, the client's objective shifts to learning model parameters that aim to maximize the loss function, as opposed to the conventional goal of minimizing it. This counterintuitive strategy is key to effectively 'forgetting' or removing the influence of the target client's data from the model.
    
    To maximize the loss function, the client undertakes multiple local iterations using stochastic gradient ascent. However, this approach of maximizing loss has its own set of challenges, particularly due to the nature of the cross-entropy loss, which is unbounded. In such cases, each step of gradient ascent tends to push the model towards increasing its loss value. The core issue here is that an unbounded increase in loss can lead the model towards a state where it effectively becomes random and loses any meaningful learning it had previously achieved. This happens because the model is being driven to continually worsen its performance with respect to the training data, rather than improving or refining its predictive capabilities. This outcome severely diminishes the utility of the model, rendering it unfeasible to obtain the unlearned model solely through gradient ascent.
    
    To address the issue of unboundedness, we introduce a constrained model to help solve the maximization problem. We first introduce a backdoor trigger in the target client on a selected party ($x^{B}_{i}$), ensuring the data on the unselected party remains clean (devoid of backdoors). To maintain the utility of the unlearned model, it is crucial to ensure that it stays reasonably close to a predetermined baseline, known as the constrained model. The constrained model acts as a reference point, representing a state where the model has effectively learned from the data distributions of the clients who have not requested unlearning (referred to as "clean" clients). Our approach to defining this constrained model involves averaging the models of all the clean clients. By doing so, we create a baseline model that reflects the collective learning from these clients' data. This average model serves as a standard or benchmark against which the unlearned model can be compared and adjusted. Considering the target client $i$ has only half of its data compromised, we preserve half of its model as the constraint. Therefore, after averaging the models of all the clean clients, $\mathbf{W}_{\text{con}} = \frac{1}{N-\frac{1}{2}}(N\mathbf{W}^{(k)}-\frac{1}{2}\mathbf{W}_{i}^{(k-1)})$. 
    After simplification, the constrained model for the target client $i$ is defined as follows:
    \begin{equation}
       \mathbf{W}_{\text{con}}=\frac{1}{2N-1}(2N\mathbf{W}^{(k)}-\mathbf{W}_{i}^{(k-1)}) 
    \end{equation}
    , where $N$ represents the total number of clients participating in VFL, and $k$ denotes the number of iterations. $\mathbf{W}^{(k)}$ signifies the weight of the global model at the $k$th round, while $\mathbf{W}_{i}^{(k-1)}$ denotes the weight of the target client $i$ at the $(k-1)$th round. It's important to note that the constrained model serves solely as a guideline for model unlearning. Being an average of all models from clean clients, it does not inherently possess the capability to defend against backdoor attacks. Subsequently, the target client $i$ optimizes over the model parameters situated within a $\ell_2$-norm ball of radius $R$ centered at $\mathbf{W}_{\text {con}}$. The radius $R$ is a hyperparameter. Hence, during the unlearning phase, the client solves the following optimization problem:
    \begin{equation}
    \max _{\mathbf{W}_{i} \in\left\{\mathbf{Z}:\left\|\mathbf{Z}-\mathbf{W}_{\text {con}}\right\|_2 \leq R\right\}}L(\mathbf{W}_{i};x^{B}_{i},y_i)
    \end{equation}
    where $L(\mathbf{W}_{i})$ is as defined in Equation \ref{lossclient}, and $\mathbf{Z}$ is a vector belonging to $\mathbb{R}^d$. By utilizing this $\ell_2$-norm ball with a radius of $R$, we ensure that the unlearned model remains closely aligned with the constrained model, thereby preserving the utility of the unlearned model. By anchoring the unlearned model to this constrained model, we aim to strike a balance: the model should diverge enough from the original to effectively unlearn the specific client's data, but not so much that it veers into randomness or loses the valuable insights gleaned from the remaining clients' data. 

    Given that this is no longer an unbounded maximization problem, we can employ gradient ascent to determine the optimized $\mathbf{W}_{i}$. Thus, we can update the parameters of the unlearned model as follows: 
    \begin{equation}
        \mathbf{W}_{i}^{(k)}\leftarrow\mathbf{W}_{i}^{(k-1)}+\eta_{u}\nabla L(\mathbf{W}_{i}^{(k-1)};x^{B}_{i},y_i)
    \end{equation}
    where $\eta_u$ is the learning rate of the unlearning model, and $(x^{B}_{i},y_i)$ is the training data sample of the target client $i$ on the selected party $B$.

    To avoid the development of a model that no longer represents meaningful learning, we incorporate an early stopping mechanism during the unlearning process. This strategy involves monitoring the distance between the gradient of the unlearning model and the gradient specific to the target client $i$ and halting the training process if the distance falls below a pre-set threshold $T$. This ensures that the unlearning process does not deteriorate the model's utility beyond a certain acceptable limit. During this unlearning phase, the server (coordinator) actively removes the contributions of the target client's local model from the global VFL model.
    
    Following the unlearning conducted by client $i$ on the selected party, it is plausible that the resultant model exhibits suboptimal performance on the data distributions of other clients. To improve the performance of the unlearned model in VFL, we propose conducting additional rounds of VFL training while specifically excluding the participation of the target client $i$ from the selected party. 
    
    \begin{algorithm}[ht]
        \SetAlgoLined
        \SetKwBlock{Begin}{Begin}{End}
        \Begin{
        \KwIn {The batch size $m$, the number of global training epochs $E$, the learning rate $\eta$}
        \textbf{Server-side:} \\
		Initialize $\mathbf{W}^{(0)}$ at random\; 
        Send $\mathbf{W}^{(0)}$ to each client\; 
		\For{$k$ from 1 to $E$}{
			Receive $\mathbf{W}^{(k)}_{i,A},\mathbf{W}^{(k)}_{i,B}$ from each client $i \in\{1, \ldots, N\}$\;
            Concatenate and compute $\mathbf{W}^{(k)}_i = \mathbf{W}^{(k)}_{i,A} \oplus \mathbf{W}^{(k)}_{i,B}$\;
            Aggregate ${\mathbf{W}}^{(k)} \leftarrow \frac{1}{N}\sum_{i=1}^N {\mathbf{W}}^{(k)}_{i}$\;
            Send $\mathbf{W}^{(k)}$ back to each client\;    
        }     
        \textbf{Client-side:} \\
        \For{$k$ from 1 to $E$}{
        Receive $\mathbf{W}^{(k-1)}$ from server\;
        \For{party $A$ and $B$ in parallel}{
            Client $i$ updates weight vector through SGD as $\mathbf{W}_{i,A}^{(k)} \leftarrow \operatorname{SGD}_A\left(\mathbf{W}^{(k-1)}\right)$, $\mathbf{W}_{i,B}^{(k)} \leftarrow \operatorname{SGD}_B\left(\mathbf{W}^{(k-1)}\right)$ \; 
            Send $\mathbf{W}_{i,A}^{(k)}, \mathbf{W}_{i,B}^{(k)}$ to the server\;
        }}
        }
    \caption{Vertical Federated Learning FedAvg}
    \label{Algo.VerFedAvg}
	\end{algorithm}
 
The procedure for training the model is outlined in Algorithm \ref{Algo.VerFedAvg}. The input for the algorithm includes the local datasets of both the target clients and all other clients. Throughout the training phase, each client applies models trained on loss using FedAvg\cite{mcmahan2017communication} within the context of vertical federated learning. In this paper, we adopt a SplitNN architecture\cite{gupta2018distributed} as the framework for VFL, as depicted in Fig~\ref{SplitNN}. Each client in parties $A$ and $B$ updates their gradients through stochastic gradient descent (SGD), utilizing the gradient of the global model, and then forwards this information to the server. On the server side, the coordinator receives these gradients from both parties $A$ and $B$, concatenates them, and proceeds to aggregate them by averaging the gradients from each client. Subsequently, it distributes the updated global model gradient back to each client in parties $A$ and $B$.

 \begin{algorithm}[ht]
        \SetAlgoLined
        \SetKwBlock{Begin}{Begin}{End}
        \Begin{
        \KwIn {The gradient of the global model $\mathbf{W}^{(E)}$, the gradient of target client $\mathbf{W}_{i}^{(E-1)}$, the batch size of unlearning $m_u$, the number of training epochs of unlearning $E_u$, the learning rate of unlearning $\eta_{u}$, the target client $i$, the radius $R$, the threshold for early stopping $T$}
        Initialize the unlearned model as $\mathbf{W}_{\text{con}}=\frac{1}{2N-1}(2N\mathbf{W}^{(E)}-\mathbf{W}_{i}^{(E-1)})$\;
		\For{$k$ from 1 to $E_u$}{
            Client $i$ updates weight vector through gradient ascent as $\mathbf{W}_{i}^{(k)}\leftarrow\mathbf{W}_{i}^{(k-1)}+\eta_{u}\nabla L(\mathbf{W}_{i}^{(k-1)};x^{B}_{i},y_i)$ subject to ${\mathbf{W}_{i} \in\left\{\mathbf{Z}:\left\|\mathbf{Z}-\mathbf{W}_{\text{con}}\right\|_2 \leq R\right\}}$\;
			\If{$\left\|\mathbf{W}_{i}^{(k)}-\mathbf{W}_{i}^{(E-1)}\right\|_2 < T$}
            {Send ${\mathbf{W}}^{(k)}_{i}$ to the server\;}  
        }
        Send ${\mathbf{W}}^{(k)}_{i}$ to the server\;
        By conducting several rounds of post-training, we can achieve a global unlearned model, denoted as $\mathbf{W}_\text{unlearn}$, which exhibits enhanced utility\;
        }
    \caption{Vertical Federated Unlearning via Gradient Ascent}
    \label{Algo.Unlearn}
	\end{algorithm}
 
   The procedure for unlearning the model is outlined in Algorithm \ref{Algo.Unlearn}. The unlearning phase commences subsequent to the vertical FedAvg training. During this phase, the target client initiates the removal of their data's influence from the global model with respect to the selected party. Upon requesting to unlearn its contribution, the unlearning model is activated. This is achieved by discarding the weight of the poisoned data, which involves utilizing the gradient from the global model, denoted as $\mathbf{W}^{(E)}$, in conjunction with the gradient specific to the target client, represented by $\mathbf{W}_{i}^{(E-1)}$. Thereafter, the target client conducts the gradient ascent to eliminate the imprints of the selected party’s contributions. To maintain the integrity and feasibility of the process, this gradient ascent is confined within predefined boundaries. After completing these steps and concluding the unlearning process, the unlearned model is obtained. Further improvement in the global unlearned model's utility can be achieved by conducting additional rounds of VFL training in the absence of the target client's data on the selected party.
	
	\section{Experiments} \label{Experiments}
	\subsection{Experiment Setup}
    Our experiments aimed to evaluate the effectiveness of vertical federated learning using various popular datasets: MNIST\cite{lecun1998mnist}, Fashion-MNIST\cite{xiao2017fashion}, and CIFAR-10\cite{krizhevsky2009learning}.
\begin{itemize}
    \item MNIST: Consists of $60,000$ training and $10,000$ testing images. Each $28\times28$ pixel grayscale image depicts handwritten digits ($0$-$9$).
\item Fashion-MNIST: Features $60,000$ training and $10,000$ testing grayscale images, each of size $28\times28$ pixels. There are $10$ distinct classes represented.
\item CIFAR-10: Contains $60,000$ colorful images of $32\times32$ pixels across $10$ categories, with $50,000$ for training and $10,000$ for testing.
\end{itemize}
To simulate the vertical federated learning setup, each image was split into two vertical halves. Party $A$ received the left half (i.e., $28\times14$ pixels for MNIST and Fashion-MNIST, and $32\times16$ pixels for CIFAR-10), while Party $B$ received the right half.

For model training, we implemented a Convolutional Neural Network (CNN) tailored to MNIST and Fashion-MNIST. For the CIFAR-10, we opted for the AlexNet model, known for its proficiency in image classification tasks. Embracing the SplitNN architecture, convolutional layers resided on the parties' end, and the fully connected linear layers were designated to the coordinator. A comprehensive summary of the model architectures can be found in Table \ref{Table.models}.
        
    \begin{table}[ht]
    \caption{Details of model architectures}
    \label{Table.models}
    \begin{adjustbox}{width=250pt}
    \centering
    \begin{tabular}{cccc}
    \hline
    Model & Side & Number & Layer \\ \hline
    \multirow{4}{*}{CNN} & \multirow{2}{*}{Party} & 1 & Convolution ($1\times32\times3\times1$) + Max Pooling ($2\times2$) \\
                         &                        & 2 & Convolution ($32\times64\times3\times1$) + Max Pooling ($2\times2$) \\ \cline{2-4}
                         & \multirow{2}{*}{Coordinator} & 1 & Fully Connected ($2688\times128$) + ReLU \\
                         &                             & 2 & Fully Connected ($128\times10$) \\ \hline
    \multirow{8}{*}{AlexNet} & \multirow{5}{*}{Party} & 1 & Convolution ($3\times64\times3\times1$) + ReLU + Max Pooling \\
                             &                        & 2 & Convolution ($64\times192\times3\times1$) + ReLU + Max Pooling \\
                             &                        & 3 & Convolution ($192\times384\times3\times1$) + ReLU + Max Pooling \\
                             &                        & 4 & Convolution ($384\times256\times3\times1$) + ReLU + Max Pooling \\
                             &                        & 5 & Convolution ($256\times256\times3\times1$) + ReLU + Max Pooling \\ \cline{2-4}
                             & \multirow{3}{*}{Coordinator} & 1 & Fully Connected ($4096\times4096$) + Dropout + ReLU \\
                             &                             & 2 & Fully Connected ($4096\times4096$) + Dropout + ReLU \\
                             &                             & 3 & Fully Connected ($4096\times10$) \\ \hline
    \end{tabular}
    \end{adjustbox}
\end{table}

    \subsection{Evaluation Metrics}
    Regarding evaluation metrics, we use clean accuracy to assess utility. Clean accuracy refers to the model's performance when tested on a clean dataset that is free of backdoor-injected data. To evaluate unlearning performance, we introduce backdoor accuracy as a measure of the backdoor's persistence. Specifically, backdoor accuracy represents the percentage of poisoned data incorrectly classified as the attacker’s desired target label, indicating whether the backdoor has been successfully removed. Additionally, we perform a Membership Inference Attack to validate our results further.

    \begin{figure}[]
    \centering
    \subfigure[Left half]{
        \includegraphics[scale=0.2]{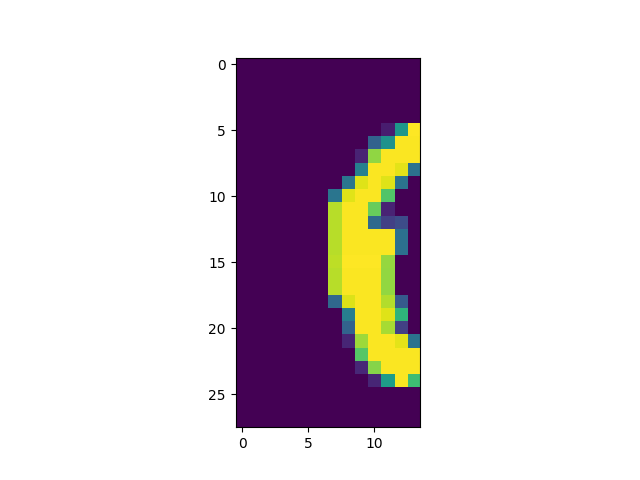}	
    }
    \subfigure[Right half]{
        \includegraphics[scale=0.2]{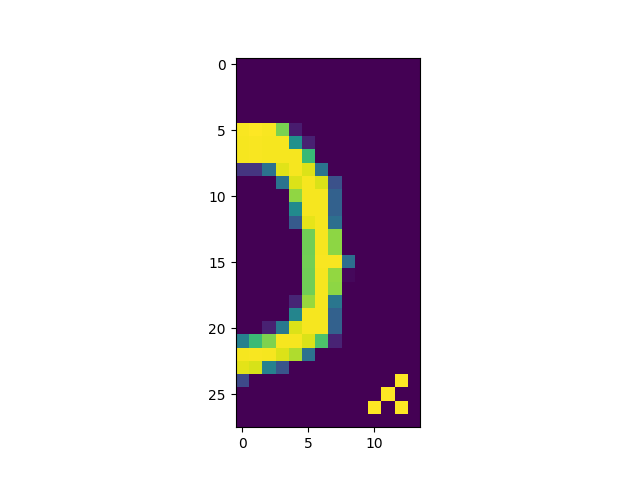}	
    }
    \caption{An example of the vertically partitioned image and the implemented backdoor in MNIST}
    \label{backdoor}
    \end{figure}

    For the backdoor trigger, we implement a $3\times3$ pixel pattern using the Adversarial Robustness Toolbox\cite{nicolae2018adversarial}, assigning the poisoned samples to the target label '$8$', as shown in Figure~\ref{backdoor}. During the backdoor attack, any data sample already labeled with the target class is excluded to prevent overlaps. As part of the evaluation, we compare our proposed method against two baseline approaches: FedAvg\cite{mcmahan2017communication}, the standard federated learning algorithm, and Retrain, where the model is retrained without the target client's participation.  
    In our experiments, we consider two distinct scenarios to evaluate the performance of our method. The first scenario includes a federated learning system with $N=5$ clients, while the second scales up to $N=10$ clients. In both cases, a significant data compromise occurs in the target client (designated as Party $B$), with up to $80\%$ of their images backdoored.

    To quantify the effect of the backdoor, we define the backdoor accuracy formula as follows:
    \begin{equation}
    Acc_{\text{backdoor}}=\frac{\sum_{i=1}^{N} \mathds{1}\left(G(x_i) = y_i\right)}{N}
    \end{equation}
    
    In this formula, $N$ represents the number of tampered samples. $G(x_i)$ is the global model's prediction for the sample $x_i$, and $y_i$ is its corresponding label. The $\mathds{1}(\cdot)$ function acts as an indicator.

    In addition to employing backdoor attacks, Membership Inference Attacks (MIA) offer a complementary perspective for validation by assessing whether individual data points can still be inferred from the model's output. This effectively examines a different dimension of data privacy and model behavior post-unlearning.
    
    For the Membership Inference Attack, we followed the methodology outlined in the study by Yeom et al. \cite{yeom2018privacy}. The first step in conducting the MIA involved constructing an attack dataset. This dataset was generated using softmax prediction vectors, $y = f_{{\text{shadow}}_i}(x)$, obtained for each data instance $x$ from the shadow models. For each distinct class $y$, a corresponding dataset $D_y$ was created. Each entry in $D_y$ consisted of a tuple: the softmax prediction vector of a class instance and a binary label. The label was set to '1' if the instance was part of the training dataset for the associated shadow model and '0' otherwise. Once the attack datasets were created, a dedicated attack model was trained for each dataset $D_y$. These models were designed to estimate the likelihood of data samples being part of the training dataset for the shadow models, using their respective softmax prediction vectors. During the testing phase, these attack models were evaluated to determine their effectiveness in inferring membership status.


   For our experiments, we employed Federated Averaging (FedAvg), retraining from scratch, and the use of a constrained model as baseline methodologies. To validate the superiority of our proposed approach, we conducted comparative analyses across three key performance indicators: accuracy on a clean dataset to assess utility, accuracy on a dataset with an embedded backdoor to evaluate the effectiveness of the unlearning approach and recall against MIA attack also to assess whether the information about target client on the selected party is removed.
       
    \subsection{Experimental Results}
    
    \subsubsection{Performance Evaluation of Baseline Methods}
    Here, we first present the accuracy results for the clean dataset. The evaluations are carried out for both $N=5$ and $N=10$ scenarios.
    \begin{figure}[]
    \centering
    \subfigure[Accuracy results on MNIST for $N=5$]{
        \includegraphics[scale=0.2]{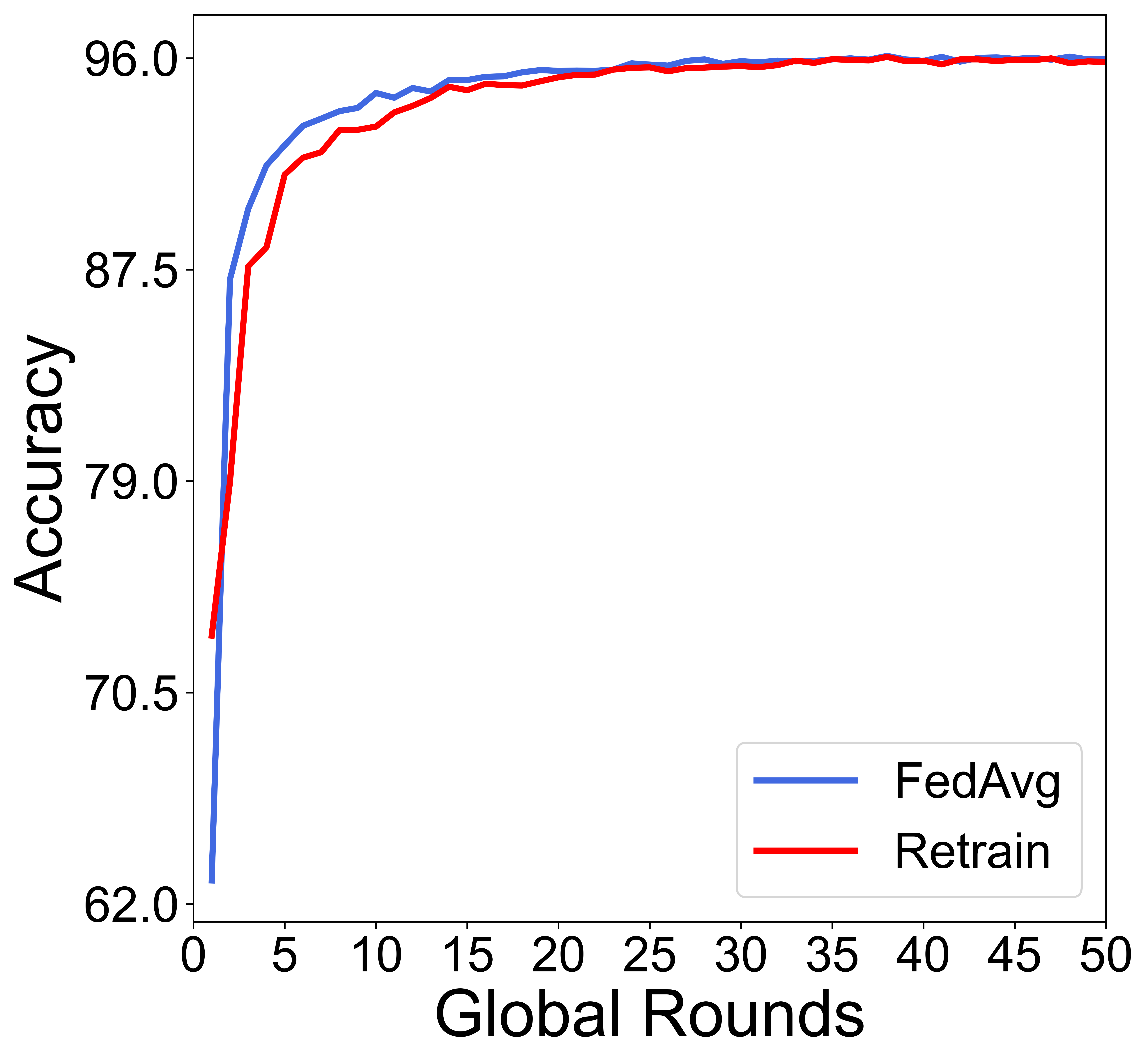}	
        \label{Fig.acc_clean_5}
    }
    \subfigure[Accuracy results on MNIST for $N=10$]{
        \includegraphics[scale=0.2]{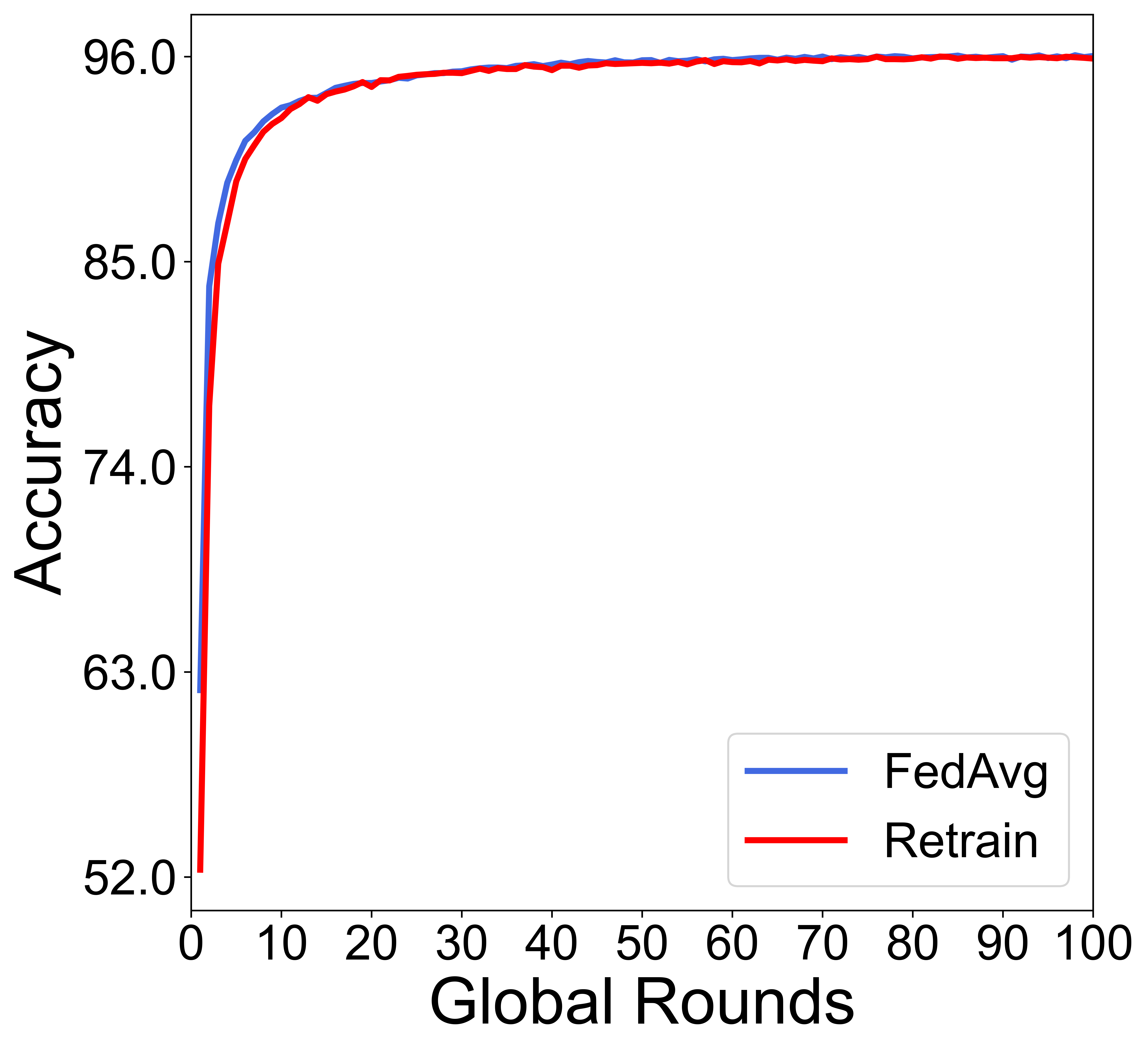}	
        \label{Fig.acc_clean_10}
    }\\
    \subfigure[Accuracy results on Fashion-MNIST for $N=5$]{
        \includegraphics[scale=0.2]{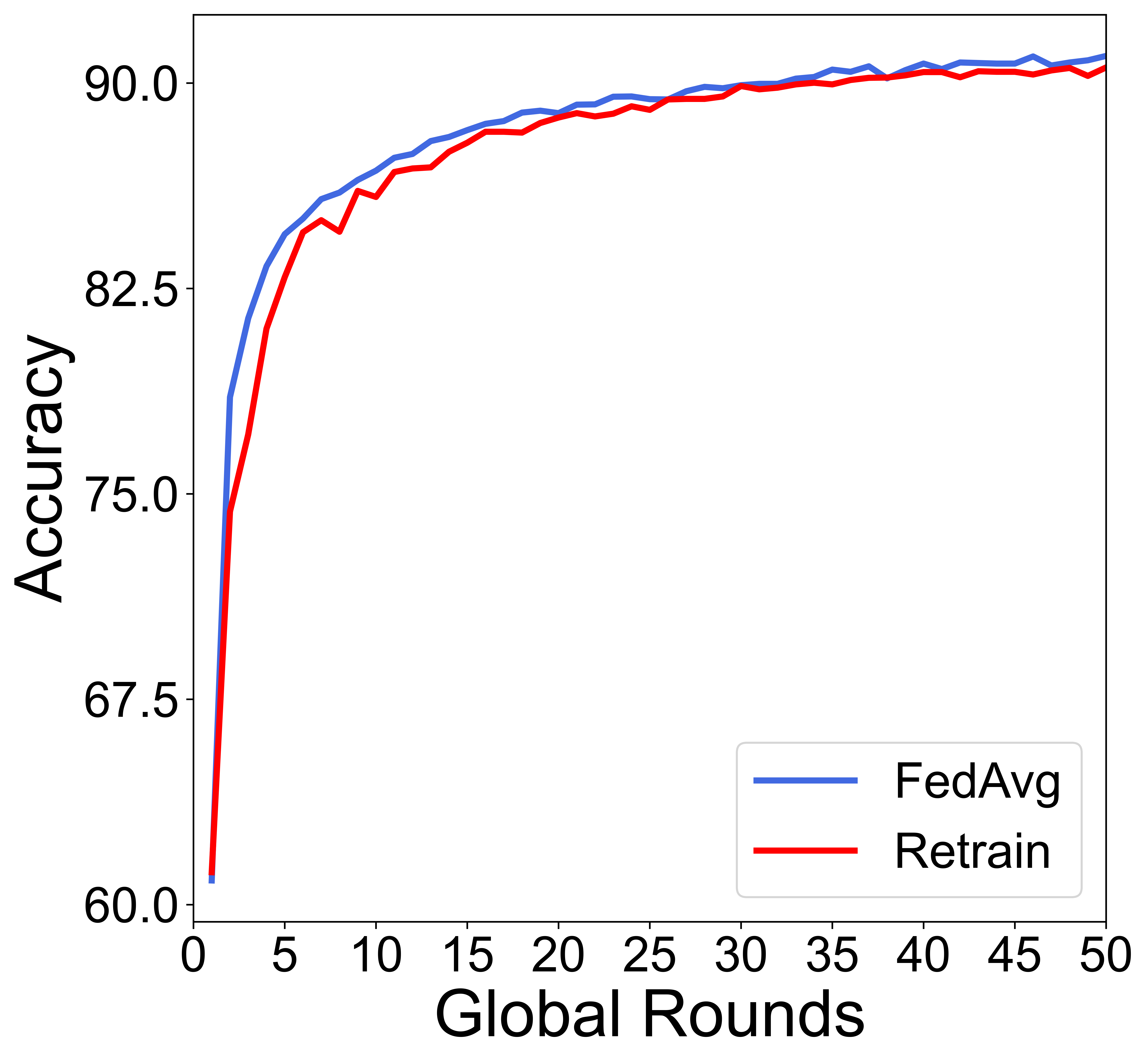}	
        \label{Fig.acc_clean_5_fashion}
    }
    \subfigure[Accuracy results on Fashion-MNIST for $N=10$]{
        \includegraphics[scale=0.2]{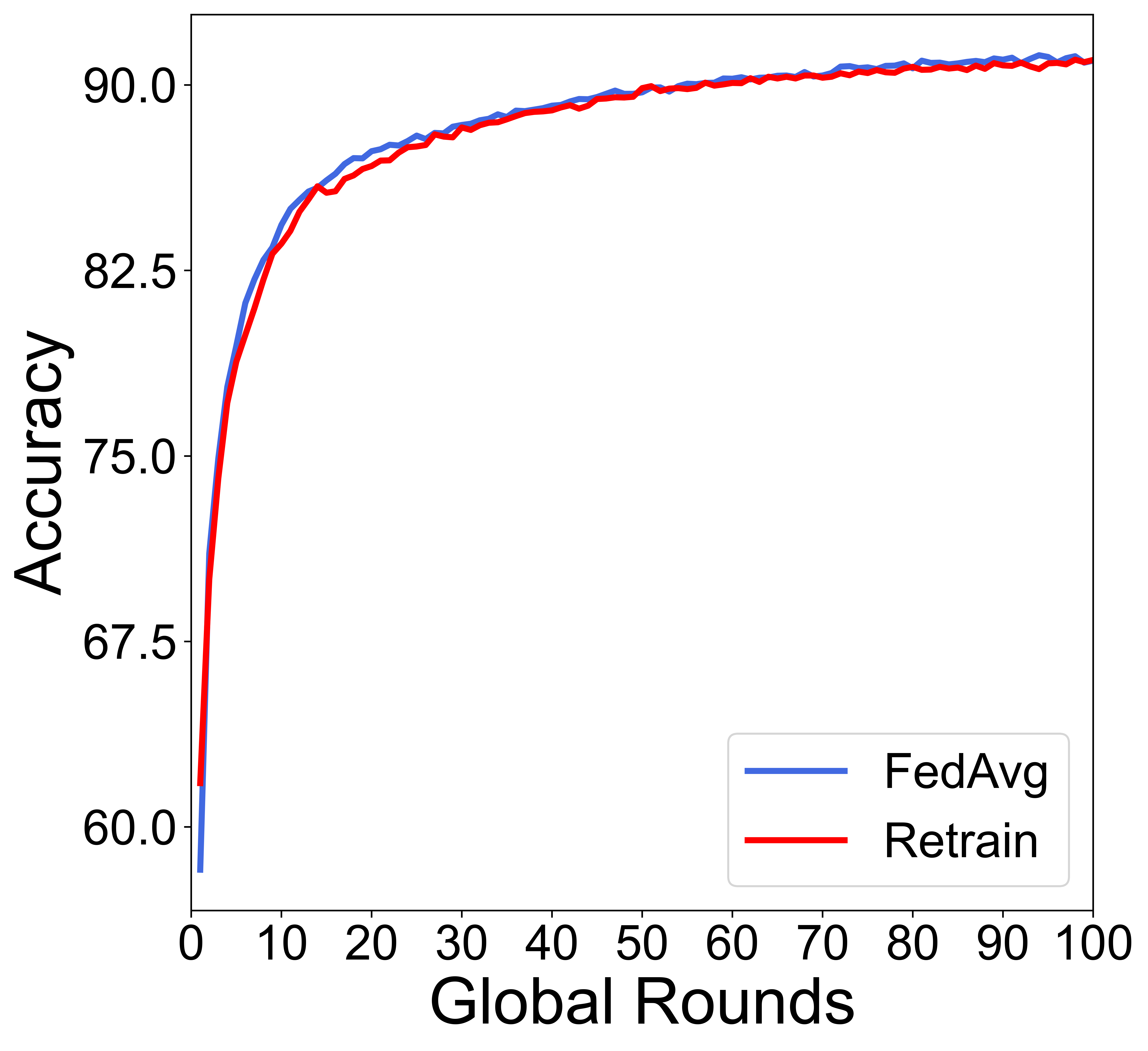}	
        \label{Fig.acc_clean_10_fashion}
    }\\
    \subfigure[Accuracy results on Cifar-10 for $N=5$]{
        \includegraphics[scale=0.2]{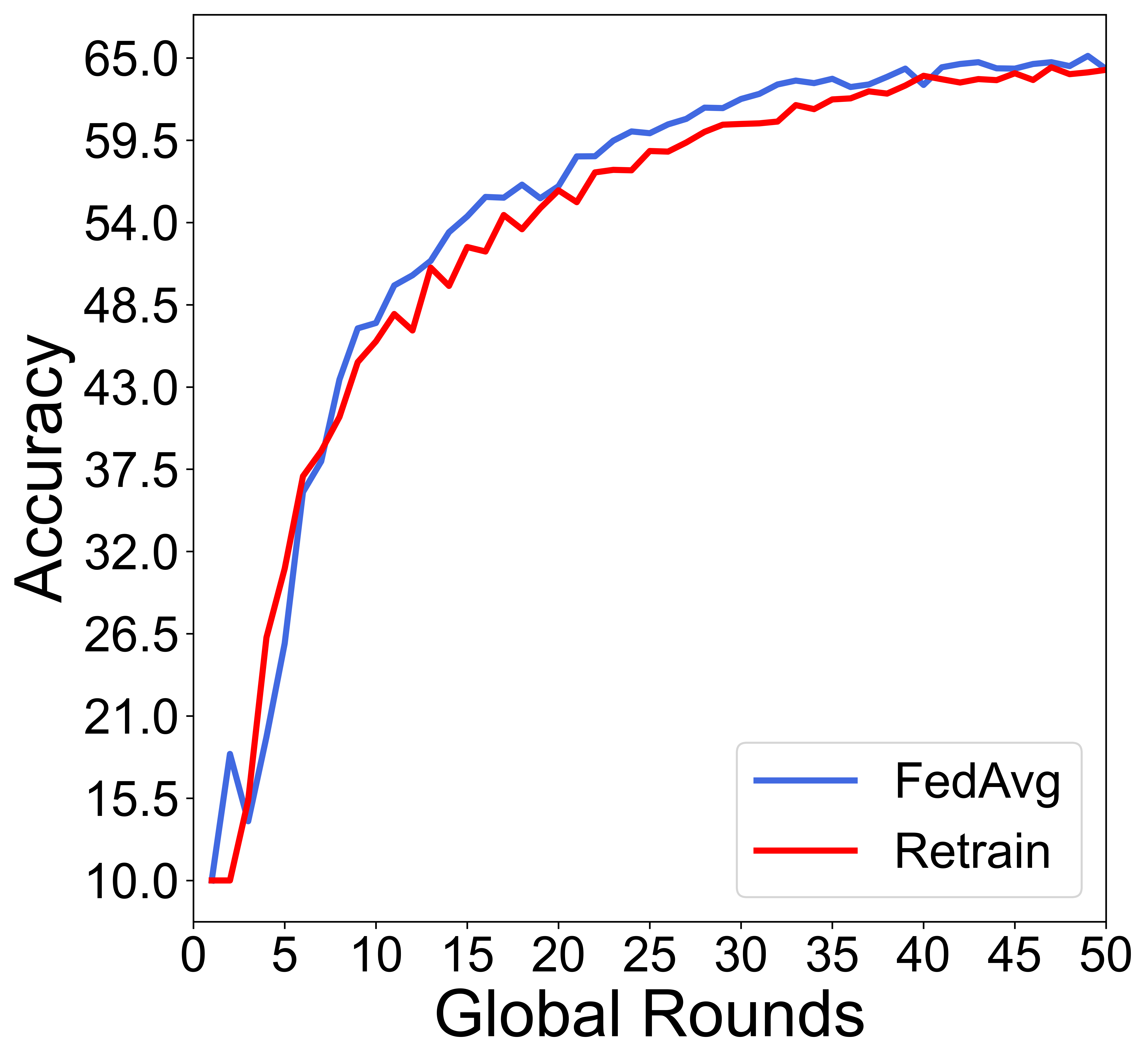}	
        \label{Fig.acc_clean_5_Cifar}
    }
    \subfigure[Accuracy results on Cifar-10 for $N=10$]{
        \includegraphics[scale=0.2]{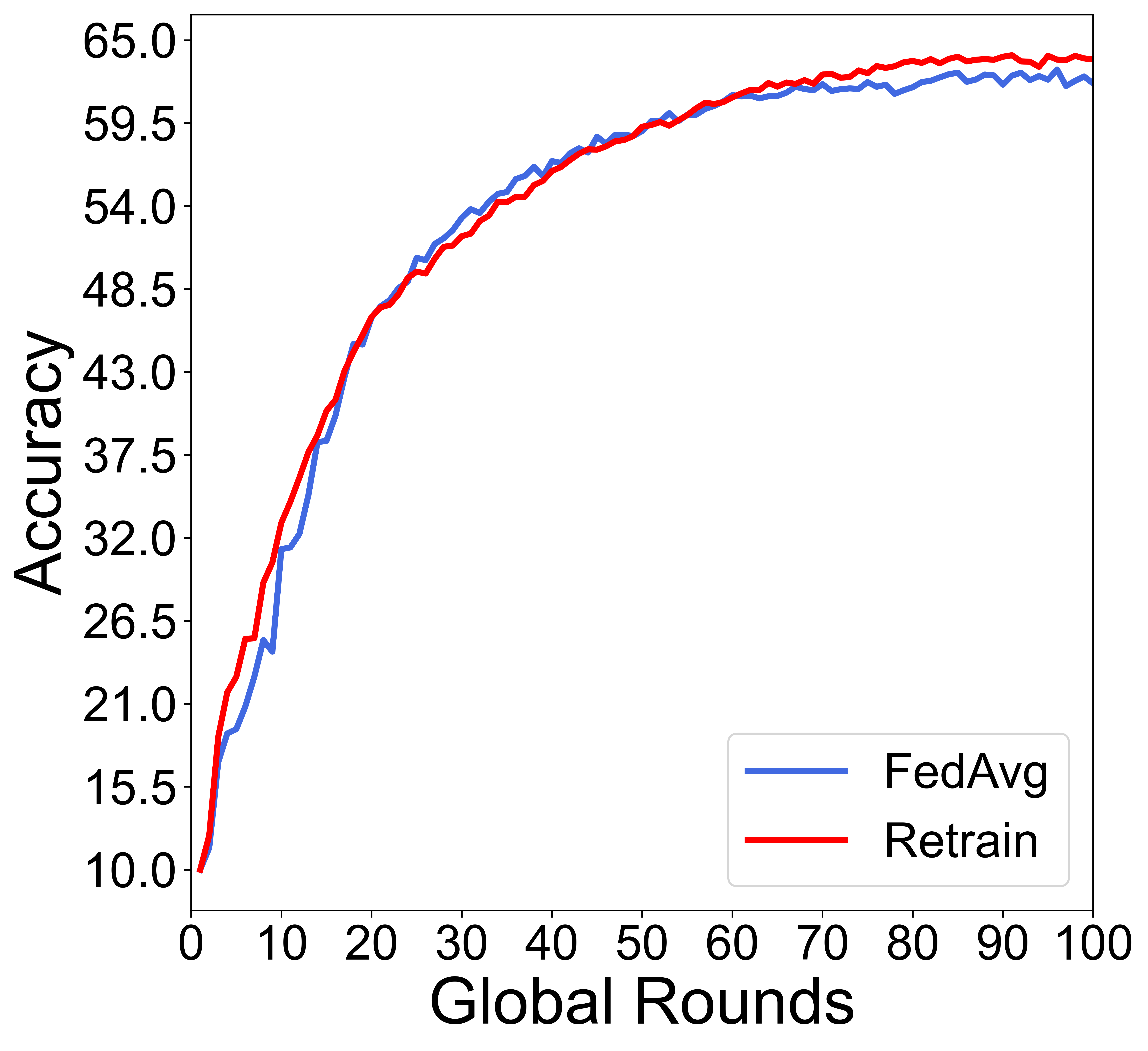}	
        \label{Fig.acc_clean_10_Cifar}
    }
    \caption{Accuracy results on the clean dataset for different datasets and values of $N$.}
    \label{Fig.acc_results_clean_combined}
\end{figure}

     Figure \ref{Fig.acc_results_clean_combined} illustrates the clean accuracy of the FedAvg and Retrain methods on the MNIST, Fashion-MNIST and Cifar-10 datasets. Given that the constrained model is derived directly from the gradient, the results of its performance are presented in Table~\ref{Table.result}. The training underwent 50 and 100 global rounds for $N=5$ and $N=10$ respectively. The accuracy increases steadily with more global epochs. There's minimal distinction between the performances of FedAvg and Retrain when starting from scratch. Both converge to high accuracy levels, showcasing optimal performance on the clean dataset irrespective of the presence of backdoor data. This observation holds true for both $N=5$ and $N=10$.
    
    \begin{figure}[ht]
    \centering
    \subfigure[Accuracy results on backdoor dataset for $N=5$]{
        \includegraphics[scale=0.2]{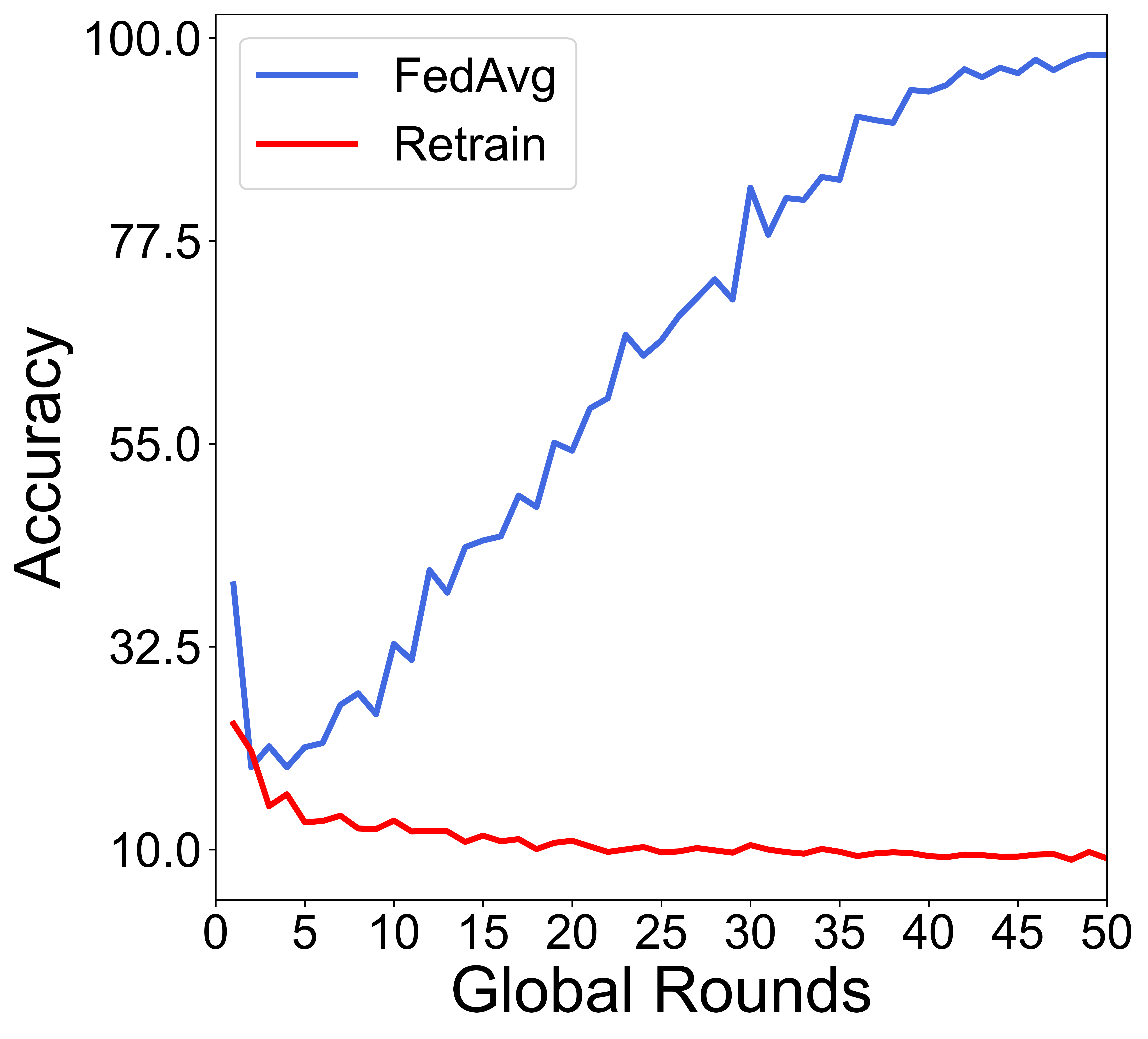}	
        \label{Fig.acc_backdoor_5}
    }
    \subfigure[Accuracy results on backdoor dataset for $N=10$]{
        \includegraphics[scale=0.2]{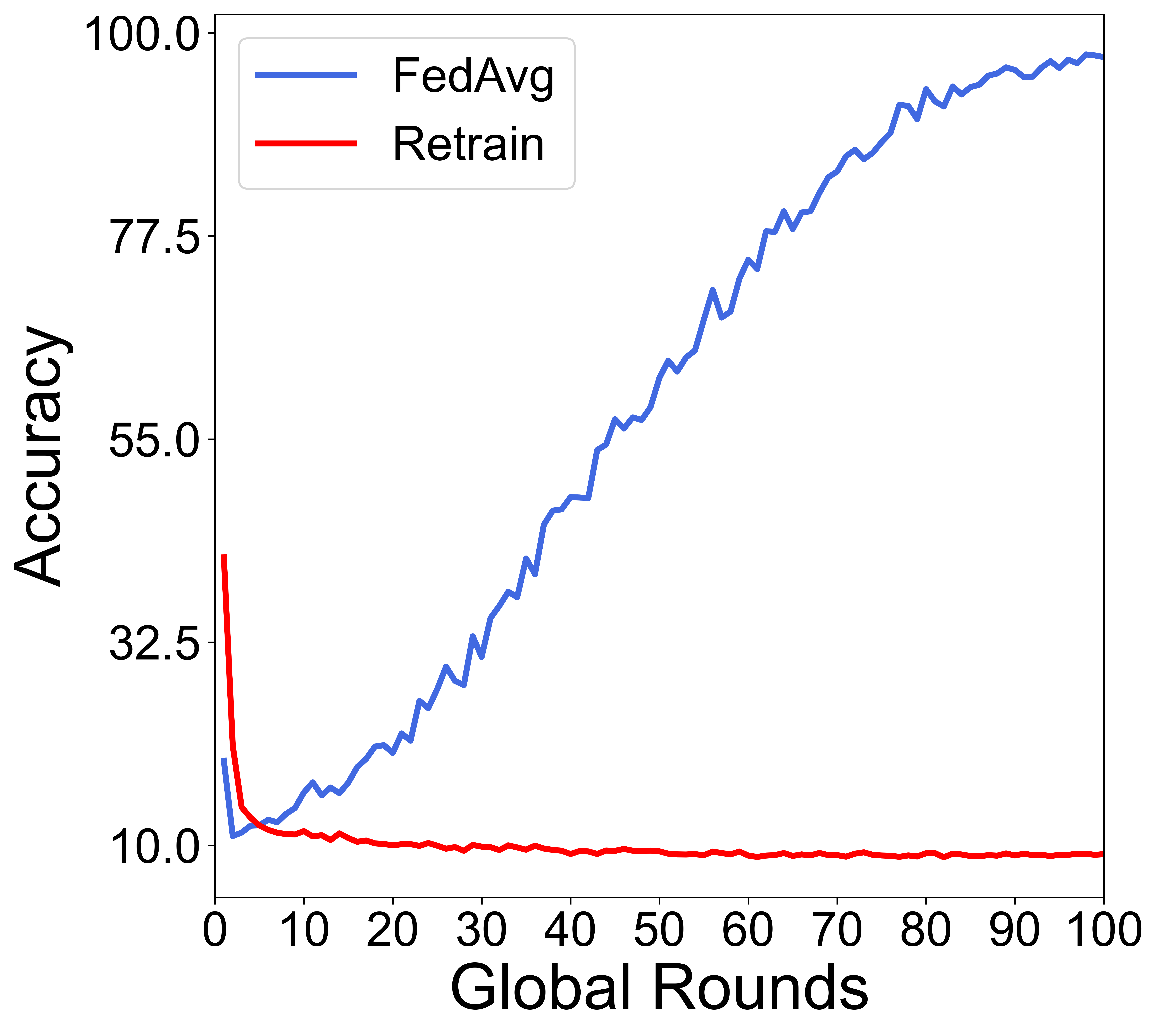}	
        \label{Fig.acc_backdoor_10}
    }\\
    \subfigure[Accuracy results on Fashion-MNIST backdoor dataset for $N=5$]{
        \includegraphics[scale=0.2]{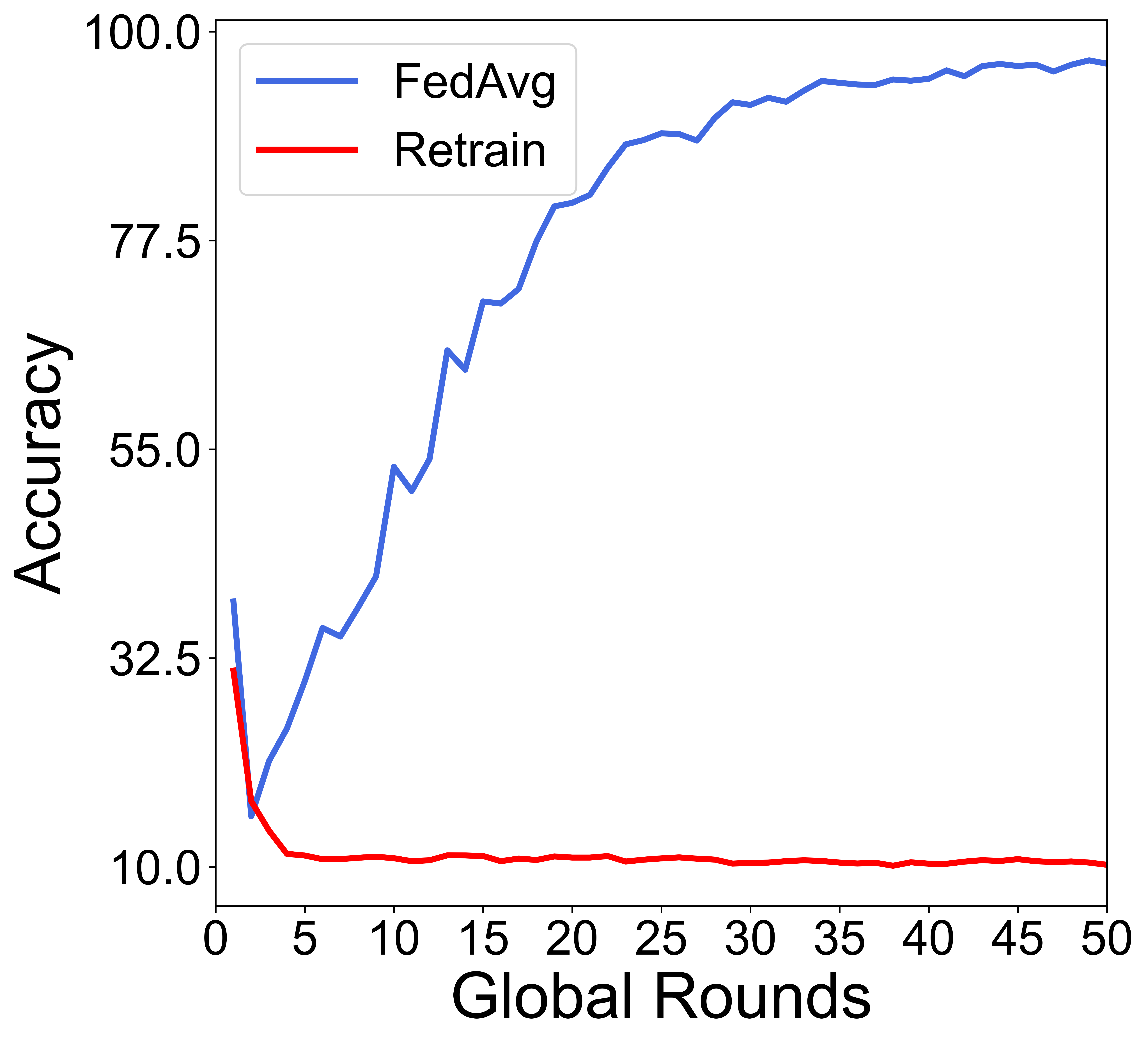}	
        \label{Fig.acc_backdoor_5_fashion}
    }
    \subfigure[Accuracy results on Fashion-MNIST backdoor dataset for $N=10$]{
        \includegraphics[scale=0.2]{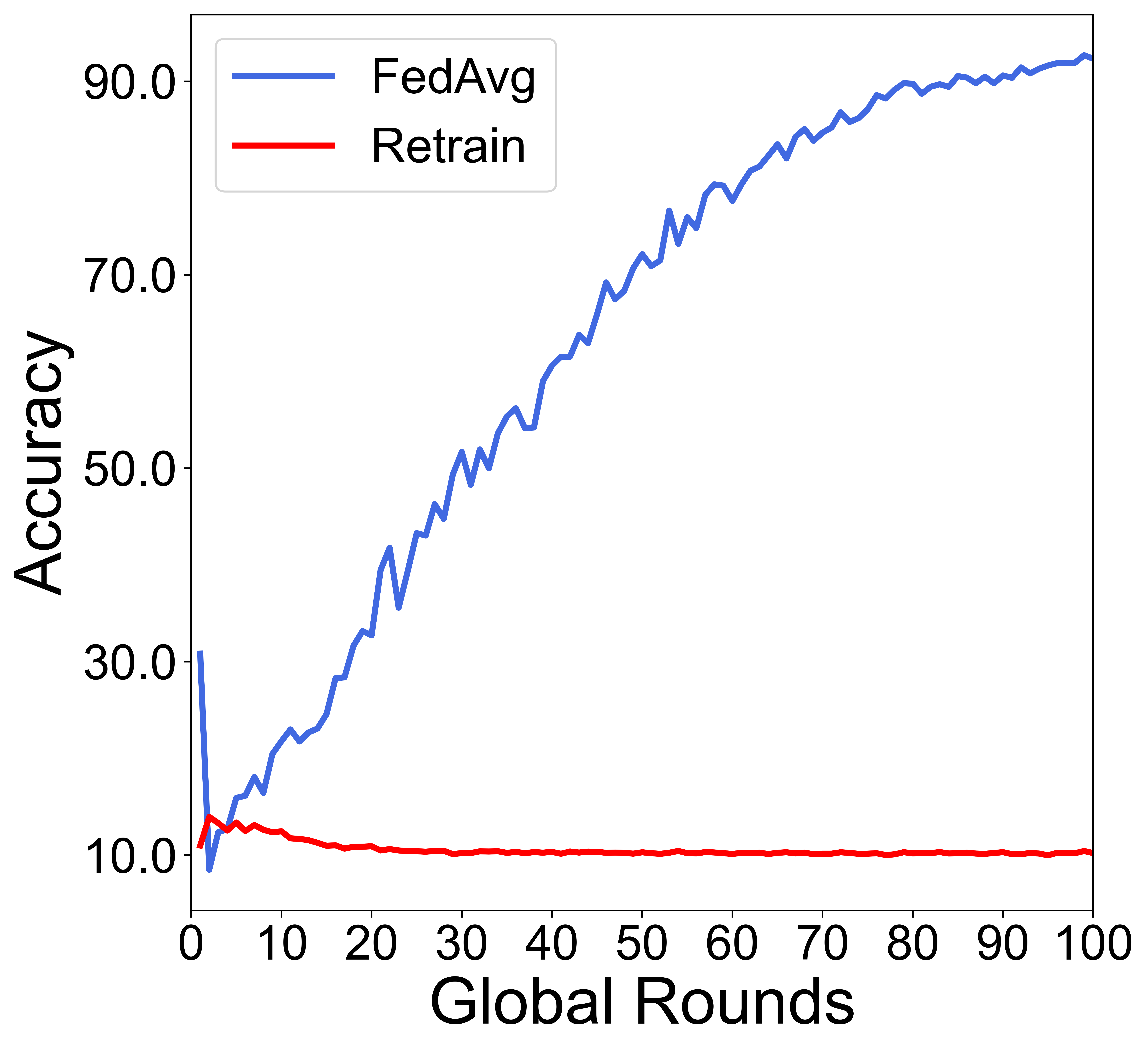}	
        \label{Fig.acc_backdoor_10_fashion}
    }\\
    \subfigure[Accuracy results on Cifar-10 backdoor dataset for $N=5$]{
        \includegraphics[scale=0.2]{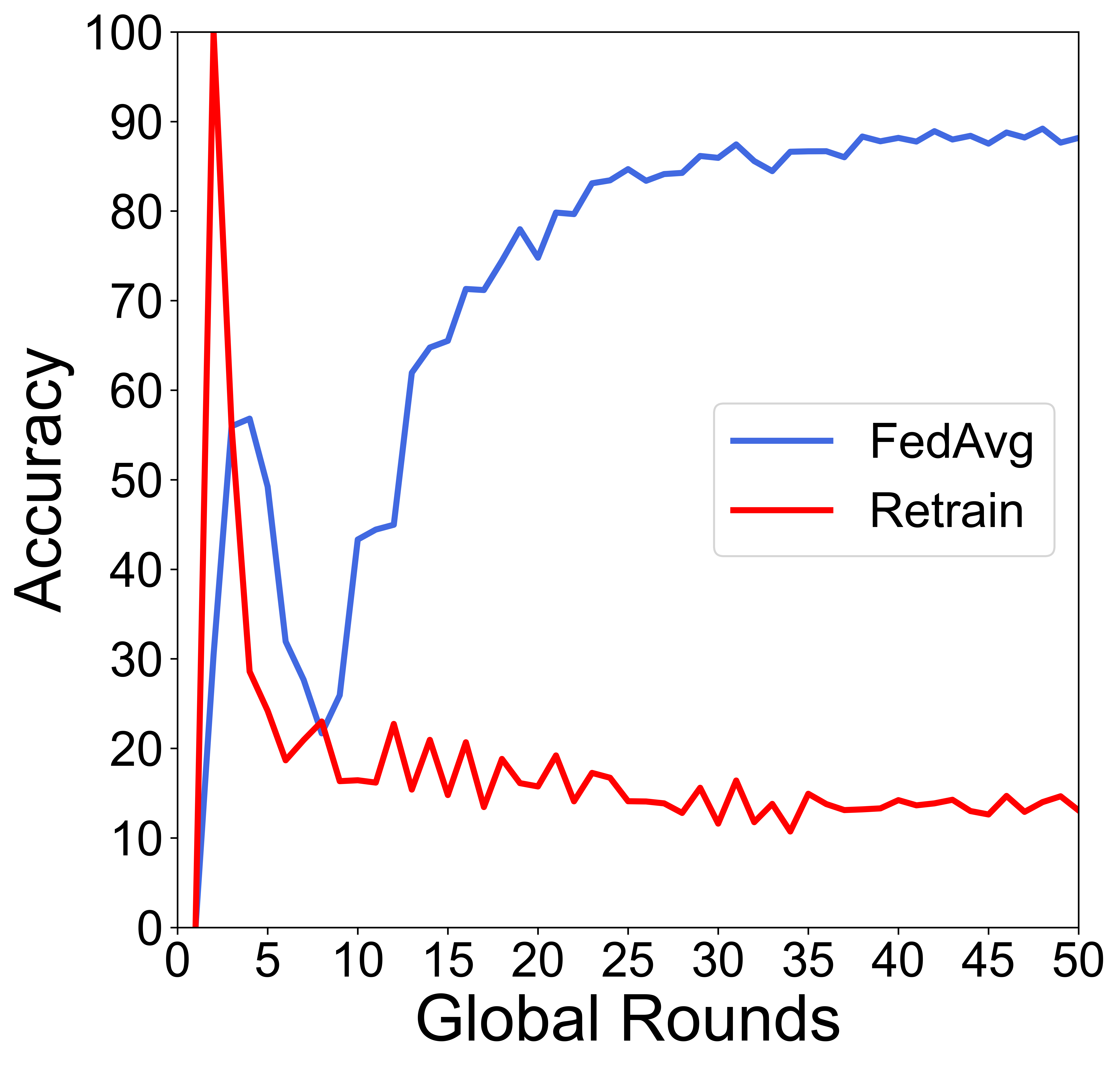}	
        \label{Fig.acc_backdoor_5_Cifar}
    }
    \subfigure[Accuracy results on Cifar-10 backdoor dataset for $N=10$]{
        \includegraphics[scale=0.2]{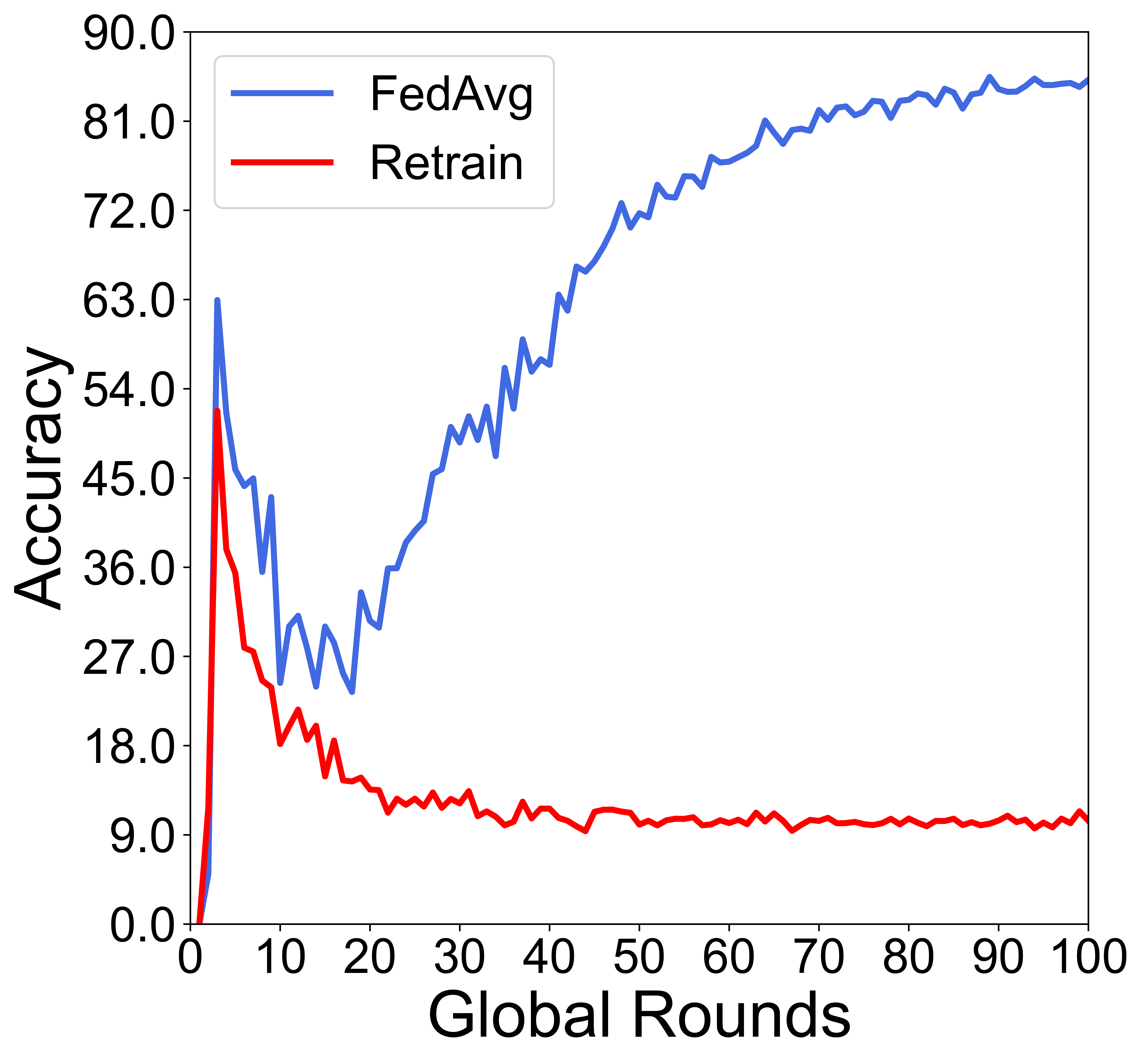}	
        \label{Fig.acc_backdoor_10_Cifar}
    }
    \caption{Accuracy results on the backdoor dataset for different datasets and values of $N$.}
    \label{Fig.acc_results_backdoor_combined}
\end{figure}

    Figure~\ref{Fig.acc_results_backdoor_combined} depicts the backdoor accuracy of FedAvg and Retrain on the MNIST, Fashion-MNIST and Cifar-10 datasets. Given that the constrained model is derived directly from the gradient, the results of its performance are presented in Table~\ref{Table.result}. The trend reveals increasing backdoor accuracy for FedAvg and decreasing for Retrain as global epochs rise. Contrary to the clean dataset results, FedAvg isn't resilient against backdoor attacks, whereas Retrain excels in this context. Since the Retrain method excludes the target client's poisoned data, backdoor triggers are ineffective. Nevertheless, a complete retrain from scratch is resource-intensive and time-consuming. The figures indicate a lengthy convergence period for the backdoor accuracy to stabilize at low levels for Retrain. The unlearning method offers a solution by eliminating the need for a complete retrain, which significantly speeds up the process. Our goal for the proposed unlearning method is to parallel the performance of the Retrain method. 

    Further details on these methods are provided in Table~\ref{Table.result} and will be elaborated upon in subsequent sections.

    \subsubsection{Performance Evaluation of Unlearning after Post-training}
    \begin{figure}[]
    \centering
    \subfigure[Clean and backdoor accuracy of unlearning methods for $N=5$ on MNIST]{
        \includegraphics[scale=0.2]{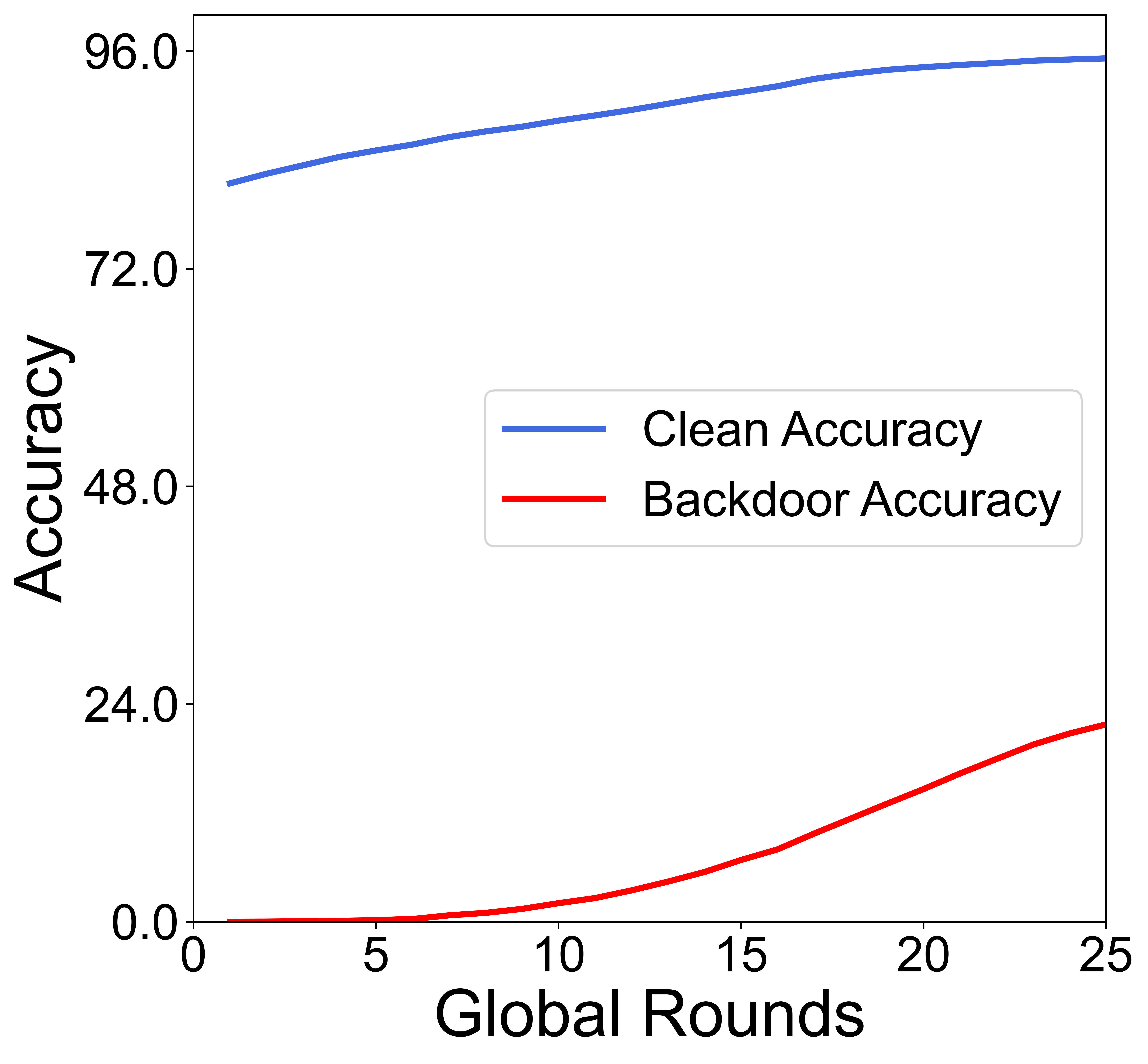}	
        \label{Fig.acc_unlearn_5}
    }
    \subfigure[Clean and backdoor accuracy of unlearning methods for $N=10$ on MNIST]{
        \includegraphics[scale=0.2]{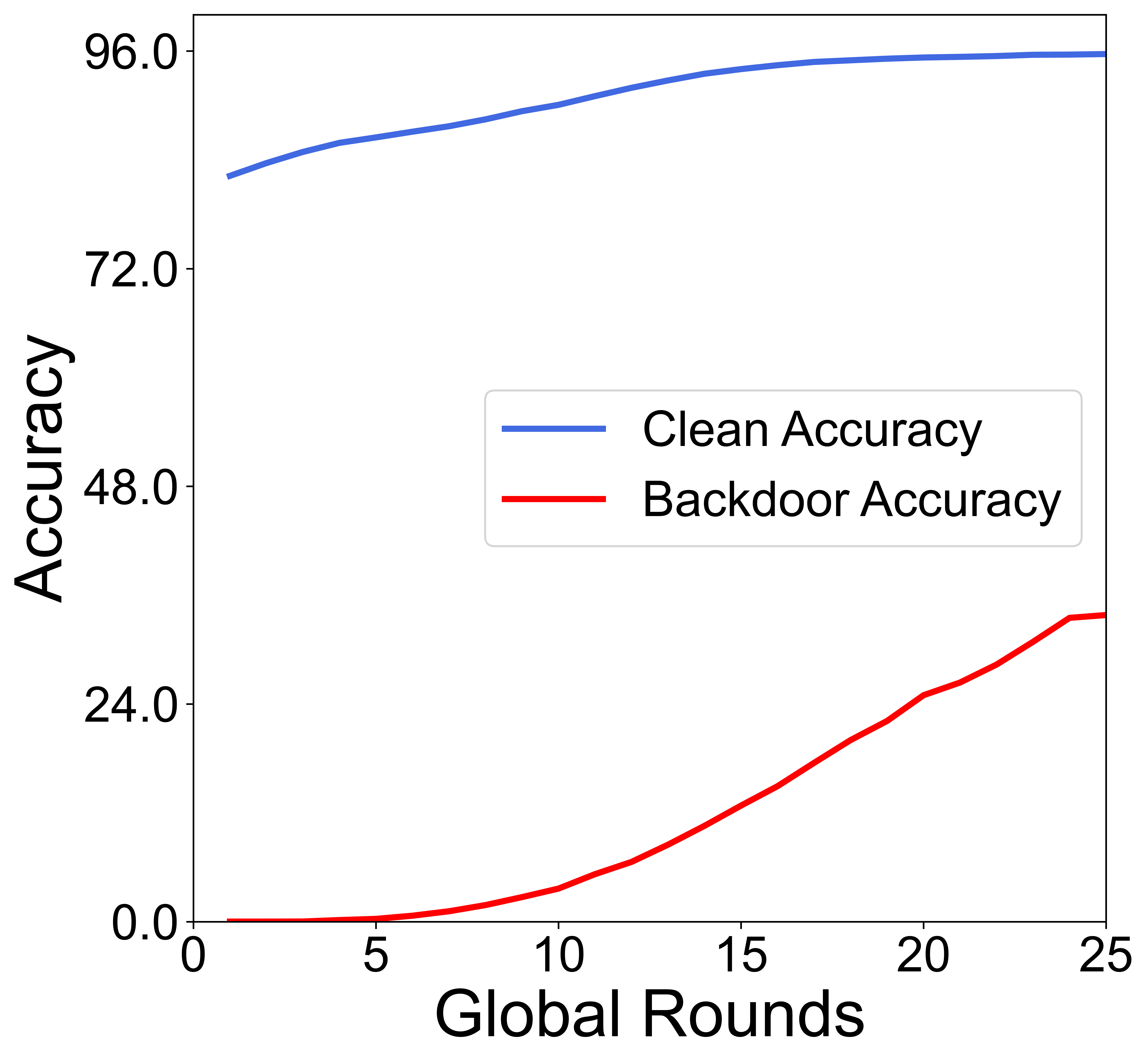}	
        \label{Fig.acc_unlearn_10}
    }\\
    \subfigure[Clean and backdoor accuracy of unlearning methods for $N=5$ on Fashion-MNIST]{
        \includegraphics[scale=0.2]{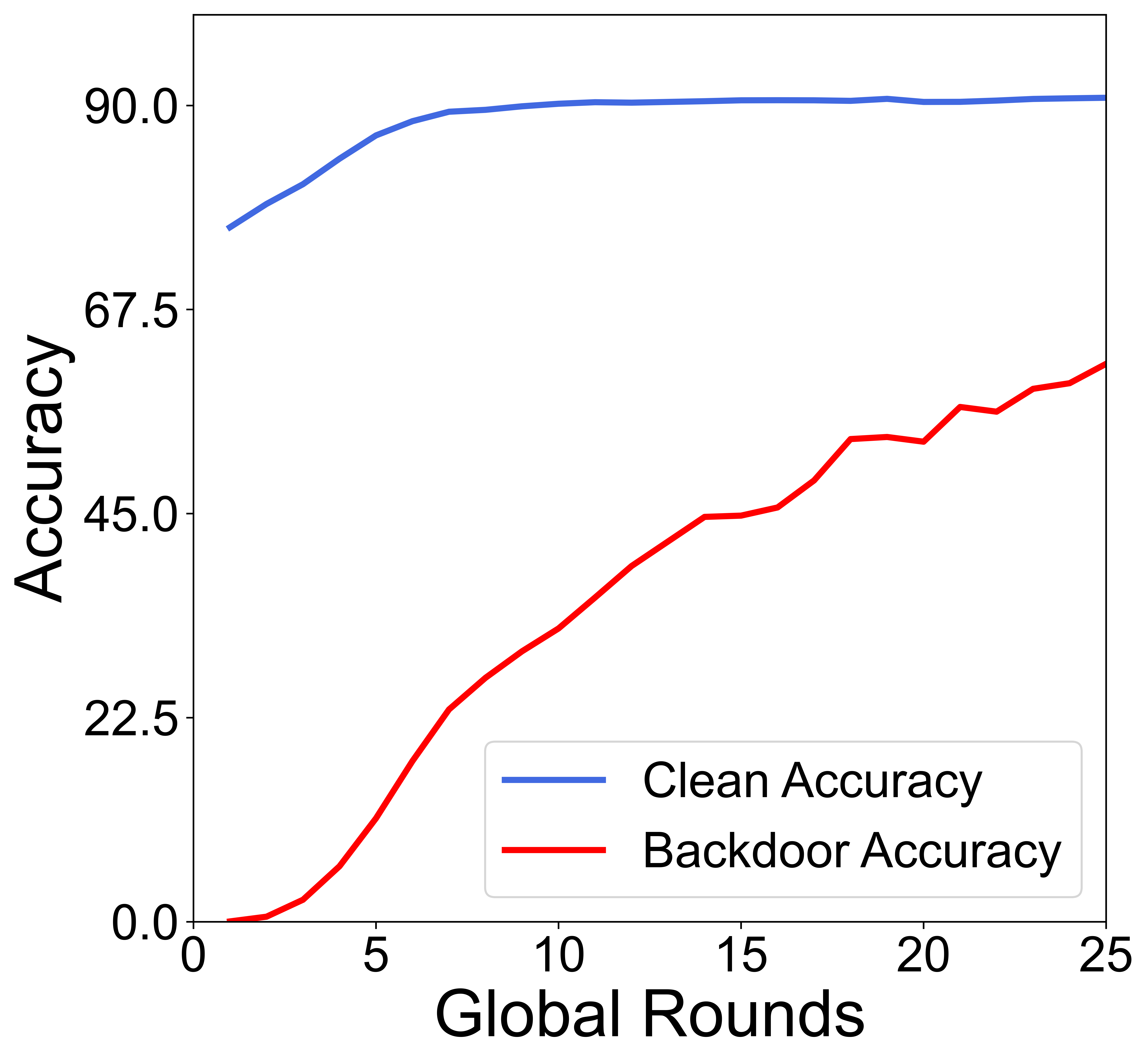}	
        \label{Fig.acc_unlearn_5_Fashion}
    }
    \subfigure[Clean and backdoor accuracy of unlearning methods for $N=10$ on Fashion-MNIST]{
        \includegraphics[scale=0.2]{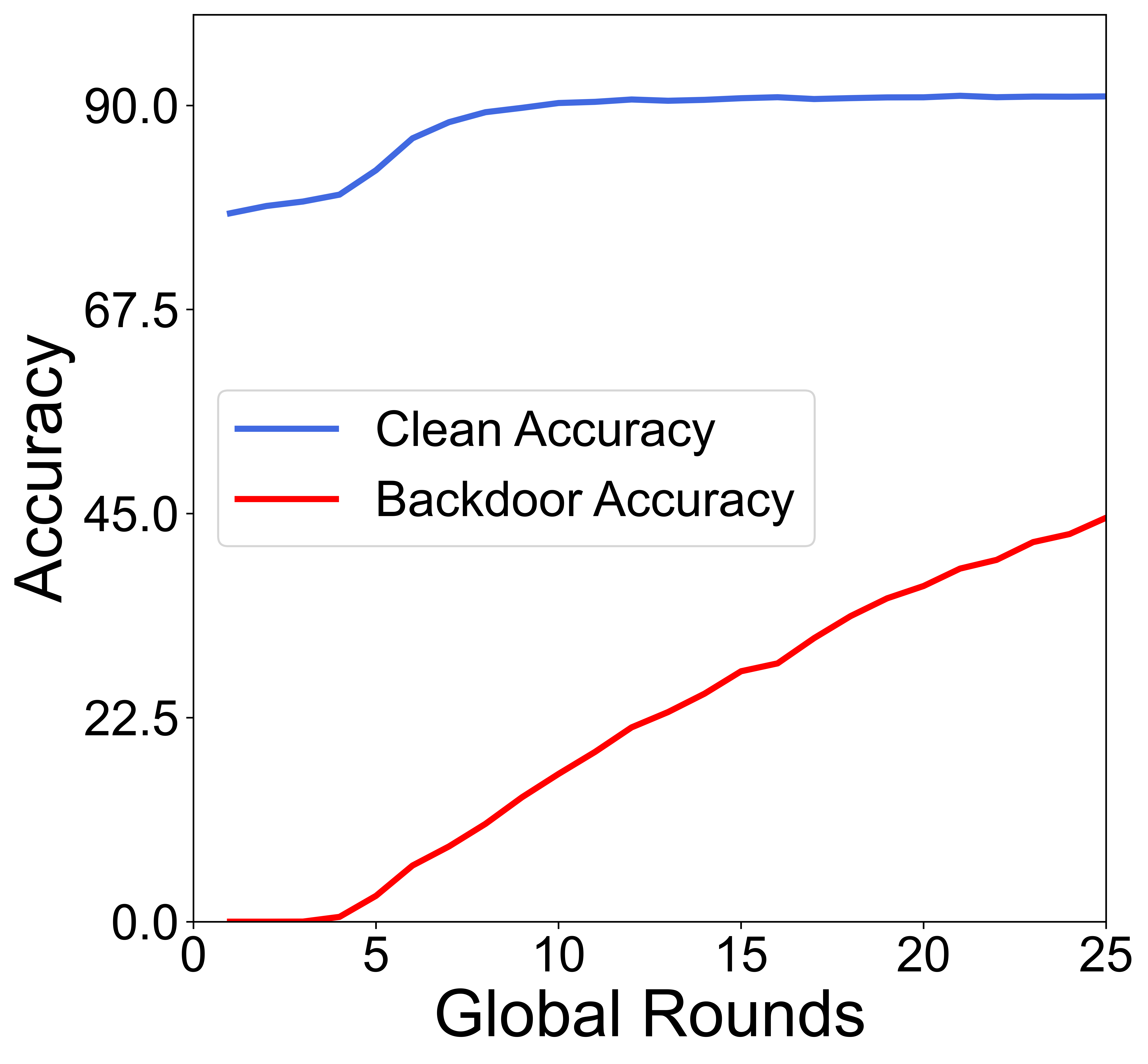}	
        \label{Fig.acc_unlearn_10_Fashion}
    }\\
    \subfigure[Clean and backdoor accuracy of unlearning methods for $N=5$ on Cifar-10]{
        \includegraphics[scale=0.2]{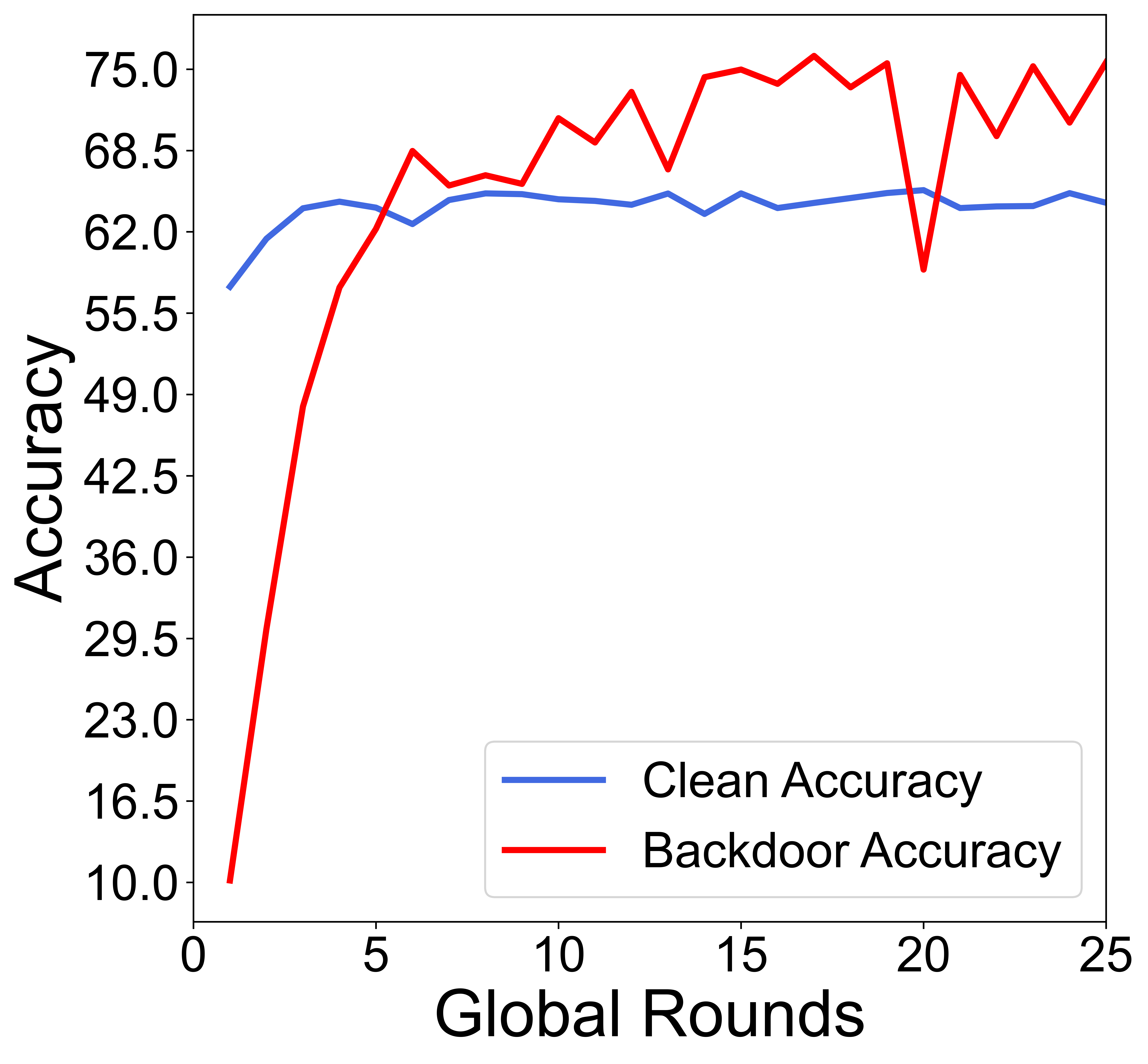}	
        \label{Fig.acc_unlearn_5_Cifar}
    }
    \subfigure[Clean and backdoor accuracy of unlearning methods for $N=10$ on Cifar-10]{
        \includegraphics[scale=0.2]{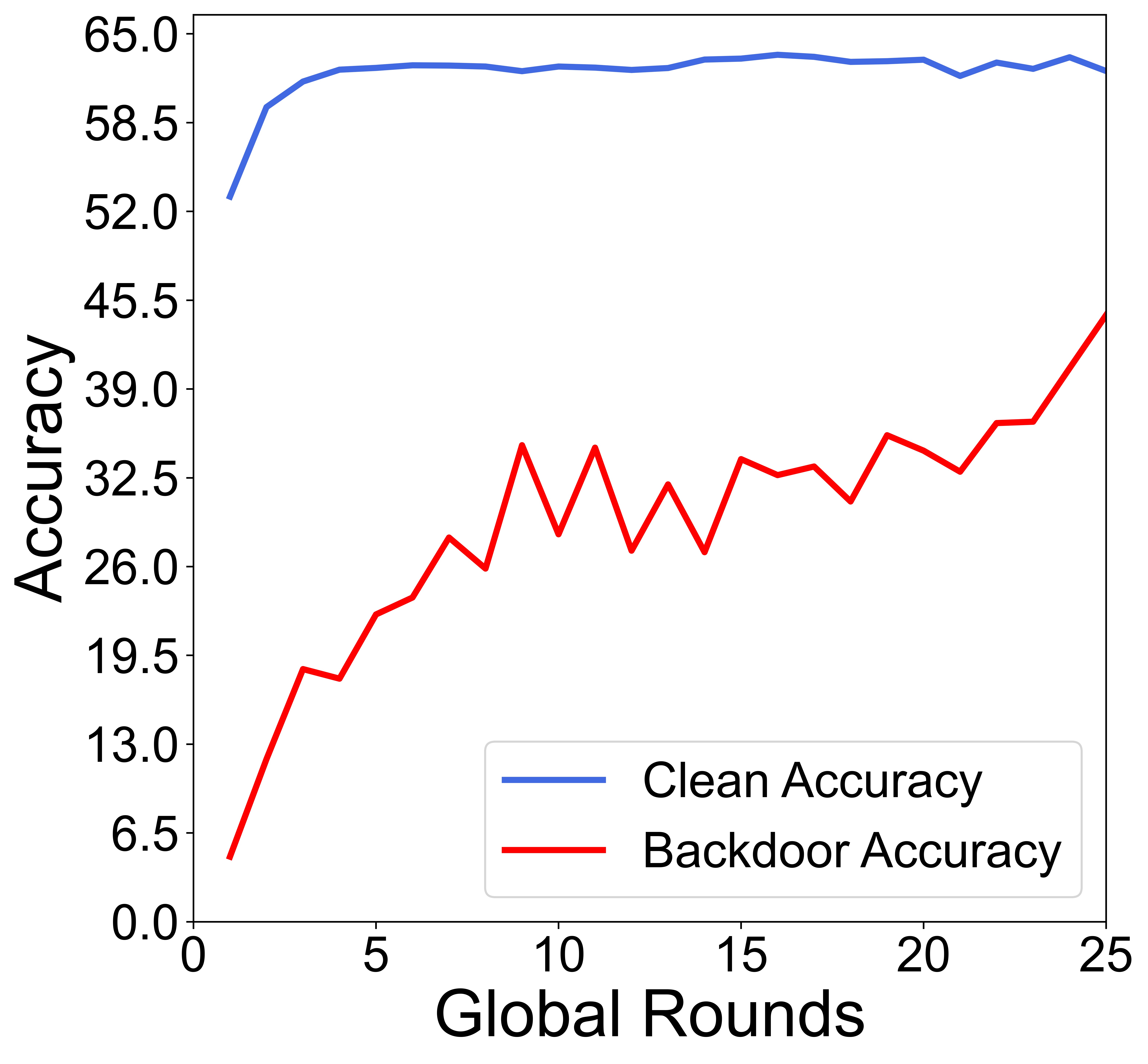}	
        \label{Fig.acc_unlearn_10_Cifar}
    }
    \caption{Clean and backdoor accuracy of unlearning methods after rounds of post-training for different datasets and values of $N$.}
    \label{Fig.acc_results_unlearn_combined}
\end{figure}
    
     Figure \ref{Fig.acc_results_unlearn_combined} illustrates the performance of our proposed unlearning method after select rounds of post-training, showcasing both clean and backdoor accuracy metrics. Remarkably, both $N=5$ and $N=10$ scenarios display comparable trends. As delineated in the methodology, the unlearning model was refined using gradient ascent, guided by the constrained model. The primary unlearning model, before any post-training, exhibits a clean accuracy of $79.84$, $80.13$, $72.25$, $77.28$, $48.25$ and $46.82$ for $N=5$ and $N=10$ on the MNIST, Fashihon-MNIST and Cifar-10 datasets, respectively, as seen in Table \ref{Table.result}. Moreover, its backdoor accuracy stands at zero, affirming that the model has effectively disregarded the tainted inputs from the target client. 
    
    However, a closer examination reveals a marginally subdued clean accuracy. To address this, we pursued continued training, excluding the poisoned data from the target client. This strategy, grounded in vertical federated learning, led to a gradual enhancement in clean accuracy with successive global epochs. In contrast, the backdoor accuracy saw a moderate uptick initially, followed by a sharper rise after a certain threshold. Given these observations, our refined approach involves limited rounds of post-training, halting after approximately $10$ global rounds for MNIST, and $1$ to $3$ rounds for Fashion-MNIST and Cifar-10. This renders our unlearning method's performance nearly on par with the Retrain method, as further elaborated in subsequent tables.
    \subsubsection{Performance Evaluation of Time Complexity}
    \begin{table*}[]
    \caption{The results for time consumption on constructing the federated model}
    \label{time}
    \begin{adjustbox}{width=500pt}
    \centering
    \begin{tabular}{|cccccc|}
    \hline
    \multicolumn{6}{|c|}{Time Consumption (second)}                                                                                                                                   \\ \hline
    \multicolumn{1}{|c|}{Dataset}       & \multicolumn{1}{c|}{FedAvg} & \multicolumn{1}{c|}{Retrain} & \multicolumn{1}{c|}{Constrained} & \multicolumn{1}{c|}{Unlearn} & Unlearn(PT) \\ \hline
    \multicolumn{1}{|c|}{MNIST}         & \multicolumn{1}{c|}{$2.81 \times 10^2$}  & \multicolumn{1}{c|}{$2.75 \times 10^2$}   & \multicolumn{1}{c|}{$3.24 \times 10^{-2}$}       & \multicolumn{1}{c|}{$2.17 \times 10^{-1}$}   & $5.53 \times 10^1$       \\ \hline
    \multicolumn{1}{|c|}{Fashion-MNIST} & \multicolumn{1}{c|}{$2.78 \times 10^2$}  & \multicolumn{1}{c|}{$2.82 \times 10^2$}   & \multicolumn{1}{c|}{$3.61 \times 10^{-2}$}       & \multicolumn{1}{c|}{$1.40 \times 10^{-1}$}   & $1.99 \times 10^1$      \\ \hline
    \multicolumn{1}{|c|}{Cifar-10}      & \multicolumn{1}{c|}{$7.05 \times 10^2$}  & \multicolumn{1}{c|}{$7.09 \times 10^2$}   & \multicolumn{1}{c|}{$9.79 \times 10^{-2}$}       & \multicolumn{1}{c|}{$9.55 \times 10^{-1}$}   & $2.67 \times 10^1$       \\ \hline
    \end{tabular}
    \end{adjustbox}
    \end{table*}
    Table~\ref{time} details the time consumption associated with our unlearning method compared to other methods in constructing global models. For this analysis, we utilized three different datasets, setting the number of clients, $N$, to 10 and the number of epochs, $E$, to 100. The FedAvg method serves as a benchmark for the performance of the final global model in vertical federated learning. The Retrain approach refers to the process of training the model anew from the ground up, specifically excluding the data from the target client associated with the selected party. In contrast, the Constrained method reflects the average efficacy of the model, specifically excluding the target client’s input for the selected party. The Unlearn method emphasizes the performance of the model after unlearning, particularly in relation to the target client and the selected party. Additionally, the Unlearn(PT) method illustrates the efficiency of the global model after undergoing several VFL training rounds without the target client's tainted data.
    
    According to our findings, the Retrain method consumes a similar amount of time as FedAvg in reconstructing the global model. In contrast, our Unlearn method significantly accelerates the global model's removal process across all three datasets. While the Constrained method is faster, being essentially a gradient operation, its effectiveness in removing contributions is limited, as detailed in Table~\ref{Table.result}. The Unlearn(PT) method requires more time than the Unlearn method due to the additional post-training rounds on a clean dataset, aimed at enhancing utility. Overall, our proposed unlearning method markedly reduces time consumption compared to retraining from scratch.
    \subsubsection{Performance Evaluation of Accuracy}     
    
    \begin{table*}[ht]
    \caption{The accuracy results for different methods on various datasets}
    \label{Table.result}
    \resizebox{21.5cm}{!}{\begin{minipage}{\textwidth}
    \begin{tabular}{cccccccccccc}
    \hline
    \multirow{2}{*}{Dataset} & Method & \multicolumn{2}{c}{FedAvg} & \multicolumn{2}{c}{Retrain} & \multicolumn{2}{c}{Constrained} & \multicolumn{2}{c}{Unlearn} & \multicolumn{2}{c}{Unlearn(PT)} \\ \cline{2-12} 
     & Clients & \multicolumn{1}{c}{N=5} & N=10 & \multicolumn{1}{c}{N=5} & N=10 & \multicolumn{1}{c}{N=5} & N=10 & \multicolumn{1}{c}{N=5} & N=10 & \multicolumn{1}{c}{N=5} & N=10 \\ \hline
    \multirow{2}{*}{MNIST} & Clean Acc & \multicolumn{1}{c}{95.98} & 96.03 & \multicolumn{1}{c}{95.85} & 95.89 & \multicolumn{1}{c}{96.01} & 96.02 & \multicolumn{1}{c}{79.84} & 80.13 & \multicolumn{1}{c}{92.11} & 92.75 \\ \cline{2-12} 
     & Backdoor Acc & \multicolumn{1}{c}{98.08} & 97.33 & \multicolumn{1}{c}{9.02} & 9.03 & \multicolumn{1}{c}{97.16} & 96.73 & \multicolumn{1}{c}{0.0} & 0.0 & \multicolumn{1}{c}{7.97} & 8.49 \\ \hline
    \multirow{2}{*}{Fashion-MNIST} & Clean Acc & \multicolumn{1}{c}{90.99} & 90.98 & \multicolumn{1}{c}{90.57} & 91.01 & \multicolumn{1}{c}{90.95} & 91.08 & \multicolumn{1}{c}{72.25} & 77.28 & \multicolumn{1}{c}{86.36} & 88.15 \\ \cline{2-12} 
     & Backdoor Acc & \multicolumn{1}{c}{96.54} & 92.33 & \multicolumn{1}{c}{10.24} & 10.21 & \multicolumn{1}{c}{95.35} & 91.22 & \multicolumn{1}{c}{0.0} & 0.0 & \multicolumn{1}{c}{8.76} & 8.30 \\ \hline
    \multirow{2}{*}{Cifar-10} & Clean Acc & \multicolumn{1}{c}{64.26} & 62.13 & \multicolumn{1}{c}{64.21} & 63.72 & \multicolumn{1}{c}{64.83} & 62.49 & \multicolumn{1}{c}{48.25} & 46.82 & \multicolumn{1}{c}{57.61} & 59.63 \\ \cline{2-12} 
     & Backdoor Acc & \multicolumn{1}{c}{88.17} & 85.15 & \multicolumn{1}{c}{13.07} & 10.42 & \multicolumn{1}{c}{86.8} & 84.0 & \multicolumn{1}{c}{0.0} & 0.0 & \multicolumn{1}{c}{10.15} & 10.9 \\ \hline
    \end{tabular}
    \end{minipage}}
    \end{table*}
 
    Table~\ref{Table.result} presents the clean and backdoor accuracy across five distinct methods for scenarios with $N=5$ and $N=10$ on three different datasets. From our observations, the proposed unlearning method adeptly expunges the target client's contribution for the selected party, securing high accuracy on clean data while reducing efficiency on backdoor data. This methodology's results align closely with the accuracy yielded by the Retrain approach. Inspecting the table reveals that FedAvg possesses both high clean and backdoor accuracy, suggesting a comprehensive understanding of both clean data and backdoor data distributions. The objective here is to unlearn the model from the backdoor data distribution. This aim is exemplified by the Retrain method, which achieves a backdoor accuracy close to $10\%$. Our unlearning strategy is anchored in the constrained model. While this model captures the clean data distribution effectively, it retains remnants of the backdoor data distribution. However, following the gradient ascent phase, this model adeptly erases the backdoor data distribution—evident from its $0\%$ backdoor accuracy—without entirely sidelining the clean data. A few subsequent rounds of FL training fine-tune this model to revive the high clean accuracy.
    
    Furthermore, a notable benefit of our VFL unlearning approach becomes apparent when juxtaposed with the Retrain method in terms of efficiency. Figures \ref{Fig.acc_results_unlearn_combined} illustrate that while the Retrain method necessitates between 30 to 40 rounds to reach a $9\%$ backdoor accuracy, the post-training phase of our unlearning method requires only about a few global rounds ($10$ for MNIST, $1$ to $3$ for Fashion-MNIST and Cifar-10) to attain similar outcomes. This methodological enhancement significantly mitigates resource consumption and time, simplifying the process to nullify the target client's impact on the selected party.
    \subsubsection{Performance Evaluation of MIA}
    \begin{table*}[]
    \caption{The results for recall of membership inference attacks}
    \label{MIA}
    \begin{adjustbox}{width=500pt}
    \centering
    \begin{tabular}{|cccccc|}
    \hline                                                                                                                  
    \multicolumn{1}{|c|}{Dataset}       & \multicolumn{1}{c|}{FedAvg} & \multicolumn{1}{c|}{Retrain} & \multicolumn{1}{c|}{Constrained} & \multicolumn{1}{c|}{Unlearn} & Unlearn(PT) \\ \hline
    \multicolumn{1}{|c|}{MNIST}         & \multicolumn{1}{c|}{$0.9899403874813711$}  & \multicolumn{1}{c|}{$0$}   & \multicolumn{1}{c|}{$0.9843040066088393$}       & \multicolumn{1}{c|}{$0$}   & $0$       \\ \hline
    \multicolumn{1}{|c|}{Fashion-MNIST} & \multicolumn{1}{c|}{$0.9922680412371134$}  & \multicolumn{1}{c|}{$0$}   & \multicolumn{1}{c|}{$0.9725130890052356$}       & \multicolumn{1}{c|}{$0$}   & $0$      \\ \hline
    \multicolumn{1}{|c|}{Cifar-10}      & \multicolumn{1}{c|}{$0.9367160775370581$}  & \multicolumn{1}{c|}{$0$}   & \multicolumn{1}{c|}{$0.9269612345678901$}       & \multicolumn{1}{c|}{$0$}   & $0$       \\ \hline
    \end{tabular}
    \end{adjustbox}
    
    \end{table*}
    
    A key objective of machine unlearning is to safeguard client privacy in federated learning. In our experiment, we employed a Membership Inference Attack (MIA) to assess the extent to which information about the unlearned client remains within the unlearned model. The MIA’s goal is to determine if a specific sample was utilized in training the original model. Consequently, a lower recall rate in MIA signifies a more effective elimination of a client’s influence from the model.

    The results, as displayed in Table~\ref{MIA}, reveal that the FedAvg method exhibits a notably high recall score. This indicates its ineffectiveness in defending membership inference attacks, as attackers can reliably recognize whether the target client's data was included in the training process. On the other hand, while the MIA attack is ineffective against the Retrain method due to its exclusion of unlearning classes, this approach is notably time-intensive, requiring complete retraining from scratch. The performance of the Constrained model is similar to FedAvg, with the MIA attack proving effective, indicating that it fails to eliminate information pertaining to the target client.
    
    In contrast, both the unlearning and unlearning (post-training) methods demonstrate robustness against MIA, suggesting that our algorithm successfully removes the target client’s impact on the selected party. Interestingly, the unlearn method’s effectiveness parallels that of the Retrain method but is achieved with significantly reduced time consumption, as evidenced in Table~\ref{time}. This experiment underscores the efficacy of our proposed unlearning algorithm in expunging a client’s influence from a trained global model.

    \subsubsection{Performance Evaluation of Parameters}
    In this section, we examine the impact of key parameters on the results, starting with the early stopping threshold $T$. This parameter is critical in preventing the unlearned model from diverging excessively, ensuring it still represents meaningful learning. To evaluate the influence of $T$, we conduct experiments on the Fashion-MNIST dataset under two scenarios: $N=5$ and $N=10$ as an example.

    Tables~\ref{T N=5} and~\ref{T N=10} reveal a clear trend: smaller values of $T$ yield higher accuracy on the clean dataset but also result in elevated backdoor accuracy. This is undesirable, as it indicates that the backdoor information embedded in the target client's poisoned data has not been effectively removed. Conversely, larger values of $T$ lead to lower backdoor accuracy, which is beneficial, but the clean accuracy significantly drops, rendering the model less useful. This occurs because the model diverges too far from the original, losing its utility in the process.

    To strike a balance, $T$ must be chosen carefully—not too small and not too large. For $N=5$, an optimal value of $T=15$ achieves a backdoor accuracy of 0, effectively eliminating the impact of poisoned data while maintaining high clean accuracy. Similarly, for $N=10$, $T=11$ achieves the same balance. At these values, the model successfully retains the clean data distribution while being entirely free of the backdoor data distribution. This ensures the utility of the model is preserved, achieving an ideal trade-off.
    
    To refine the selection of $T$, we conduct a grid search over a reasonable interval, applying this strategy to determine $T$ values for other datasets. As a result, our unlearned model captures the clean data distribution effectively while eliminating traces of backdoor information. Furthermore, this optimized unlearned model can undergo additional post-training to further enhance performance.
    \begin{table*}[]
    \caption{Performance evaluation of the threshold T on the Fashion-MNIST dataset for N=5}
    \label{T N=5}
    \begin{adjustbox}{width=500pt}
    \centering
    \begin{tabular}{|c|c|c|c|c|c|c|c|c|c|c|c|c|c|c|}
    \hline
    The Value of T & 1     & 3     & 5     & 7     & 9     & 10    & 11    & 13    & 15    & 20    & 30    & 50    & 70   & 100  \\ \hline
    Clean Acc      & 90.63 & 90.45 & 90.57 & 90.66 & 84.67 & 82.36 & 80.86 & 77.42 & 76.04 & 70.63 & 26.24 & 11.05 & 10.0 & 10.0 \\ \hline
    Backdoor Acc   & 91.36 & 90.88 & 92.31 & 91.28 & 33.63 & 14.27 & 3.08  & 0.78  & 0.0   & 0.0   & 0.0   & 0.0   & 0.0  & 0.0  \\ \hline
    \end{tabular}
    \end{adjustbox}
    \end{table*}
    
    \begin{table*}[]
    \caption{Performance evaluation of the threshold T on the Fashion-MNIST dataset for N=10}
    \label{T N=10}
    \begin{adjustbox}{width=500pt}
    \centering
    \begin{tabular}{|c|c|c|c|c|c|c|c|c|c|c|c|c|c|c|}
    \hline
    The Value of T & 1     & 3     & 5     & 7     & 9     & 10    & 11    & 13    & 15    & 20    & 30    & 50    & 70   & 100  \\ \hline
    Clean Acc      & 90.59 & 90.69 & 90.68 & 82.15 & 80.77 & 79.89 & 78.88 & 77.51 & 76.25 & 70.26 & 53.72 & 31.61 & 10.0 & 10.0 \\ \hline
    Backdoor Acc   & 79.67 & 82.14 & 81.32 & 10.89 & 2.27  & 0.12  & 0.0   & 0.0   & 0.0   & 0.0   & 0.0   & 0.0   & 0.0  & 0.0  \\ \hline
    \end{tabular}
    \end{adjustbox}
    \end{table*}
    
    \begin{table}[]
    \caption{Performance evaluation of the radius R on the Fashion-MNIST dataset for N=5}
    \label{R N=5}
    \begin{adjustbox}{width=250pt}
    \centering
    \begin{tabular}{|c|c|c|c|c|c|c|c|c|}
    \hline
    The Value of R & 5*Dist & 3*Dist & 2*Dist & Dist & Dist/2 & Dist/3 & Dist/5 & Dist/10 \\ \hline
    Clean Acc      & 77.95  & 75.63  & 74.4   & 72.1 & 74.51  & 73.85  & 68.22  & 69.28   \\ \hline
    Backdoor Acc   & 0.05   & 0.12   & 0.09   & 0.07 & 0.37   & 0.04   & 0.01   & 0.29    \\ \hline
    \end{tabular}
    \end{adjustbox}
    \end{table}
    
    \begin{table}[]
    \caption{Performance evaluation of the radius R on the Fashion-MNIST dataset for N=10}
    \label{R N=10}
    \begin{adjustbox}{width=250pt}
    \centering
    \begin{tabular}{|c|c|c|c|c|c|c|c|c|}
    \hline
    The Value of R & 5*Dist & 3*Dist & 2*Dist & Dist  & Dist/2 & Dist/3 & Dist/5 & Dist/10 \\ \hline
    Clean Acc      & 78.24  & 78.84  & 79.84  & 79.51 & 79.97  & 78.4   & 79.72  & 80.26  \\ \hline
    Backdoor Acc   & 0.06   & 0.26   & 0.19   & 0.15  & 0.12   & 0.0    & 0.13   & 0.65    \\ \hline
    \end{tabular}
    \end{adjustbox}
    \end{table}
    Next, we explore the impact of the radius $R$, which defines the $\ell_2$-norm ball around $\mathbf{W}_{\text{con}}$, plays a crucial role in controlling how far the unlearned model deviates from the constrained model. Take the Fashion-MNIST dataset as an example, we tested various radius values for $N=5$ and $N=10$. As shown in Table~\ref{R N=5} and Table~\ref{R N=10}, the results indicate only minor differences across different radius values.

    Here, $Dist$ represents the average Euclidean distance between $\mathbf{W}_{\text{con}}$ and a random model, calculated by averaging over 10 random models, with the early stop threshold activated. A larger radius allows the unlearned model to drift farther from the constrained model, which can lead to a deterioration in accuracy on the clean dataset, ultimately impacting the model's utility. Conversely, a smaller radius keeps the unlearned model too close to the constrained model, preserving the backdoor features and resulting in higher backdoor accuracy, indicating that more poisoned data remains in the model.
    
    Based on these observations, we set the radius $R$ to $Dist/3$ for all datasets. This choice strikes a balance by ensuring the model stays closer to the constrained model than a random model, while also providing enough flexibility to effectively undergo the unlearning procedure. This approach prevents the model from being overly constrained, thereby maintaining its ability to unlearn poisoned data effectively.
         
    \section{Conclusion and Future Direction} \label{Conclusion}
   In this paper, we explore the complex landscape of vertical federated unlearning, underscoring its inherent challenges, and introducing an advanced unlearning algorithm. Our approach utilizes gradient ascent within a constrained model framework, specifically targeting the client's contribution. The effectiveness of this method is substantiated by extensive experimentation on various well-known datasets. Recognizing that an unlearned model can subtly retain fragments of information from target clients is a critical aspect of the unlearning process in machine learning. While these remnants might not be immediately apparent, subsequent in-depth analysis could uncover this residual data. This realization underscores the need for ongoing research into potential information leakages in unlearned models.

   While our proposed method demonstrates promising results, several aspects require further exploration. One significant area is to apply more complex and diverse datasets, particularly those with higher feature dimensions or pronounced inter-client feature disparities. While our current study employs datasets such as MNIST, Fashion-MNIST, and CIFAR-10, which are effective for proof-of-concept, they may not fully capture the challenges inherent in more intricate VFL scenarios. Expanding the evaluation to include real-world datasets tailored to vertical federated learning could enhance the robustness and generalizability of our findings. In addition, our future investigation will focus on various data distribution, particularly the challenges posed by heterogeneous and non-IID data distributions.

	\bibliographystyle{IEEEtran}
    \bibliography{ref}
 
\begin{IEEEbiography}[{\includegraphics[scale=0.42]{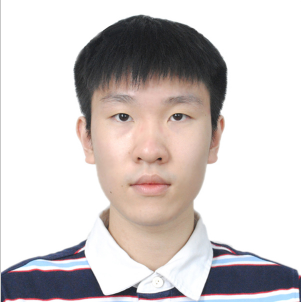}}]{Mengde Han} received the B.S. degree from Shandong University, China, in 2016 and the M.S.Eng. degree from Johns Hopkins University, USA in 2017. He is currently pursuing the Ph.D. degree with the University of Technology Sydney, Australia. His research interests are federated learning, fairness and privacy preserving.
\end{IEEEbiography}
 
\vspace{11pt}

\begin{IEEEbiography}[{\includegraphics[scale=0.8]{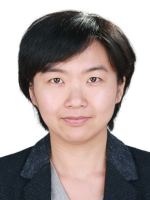}}]{Tianqing Zhu}
is a professor and associate dean of the Faculty of Data Science in City University of Macau. She received the B.Eng. degree and M.Eng. degree from Wuhan University, Wuhan, China, in 2000 and 2004, respectively, and the Ph.D. degree from Deakin University, Australia, in 2014. She was an associate professor in the School of Computer Science, University of Technology Sydney, Australia, from 2020 to 2024. Her research interests include privacy preserving, cyber security and privacy in the artificial intelligence.
\end{IEEEbiography}

\vspace{11pt}

\begin{IEEEbiography}[{\includegraphics[scale=0.45]{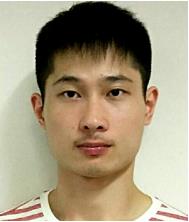}}]{Lefeng Zhang} received the B.Eng. and M.Eng. degrees from the Zhongnan University of Economics and Law, China, in 2016 and 2019, respectively and his Ph.D. degree fromUniversity of Technology Sydney, Australia. He is currently an assistant professor at City University of Macau, Macao. His research interests are game theory and privacy preserving.
\end{IEEEbiography}

\vspace{11pt}

\begin{IEEEbiography}[{\includegraphics[scale=0.12]{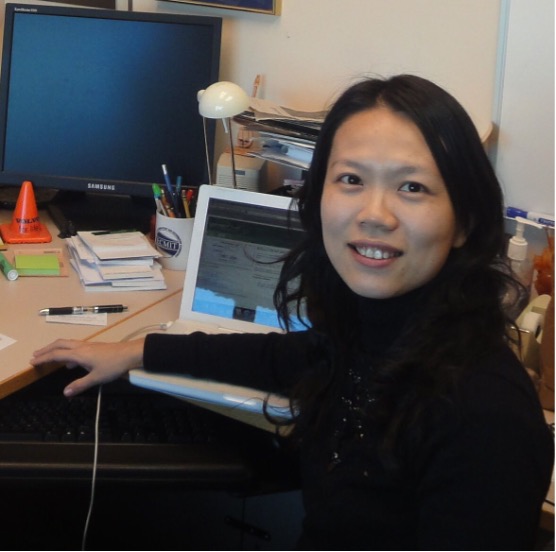}}]{Huan Huo}
received the B.Eng and Ph.D. degrees from Northeastern University, China in 2002 and 2007, both in Computer Science and Technology. From 2012 to 2014, Huan Huo taught at the Department of Computer Information System, the University of the Fraser Valley in Canada, and did collaborative research in the University of Waterloo as a visiting scholar for one year. Since 2018, she has been a senior lecture in the school of software at the University of Technology Sydney, Australia. Her research interests include data stream management technology, advanced data analysis, and data-driven cybersecurity.
\end{IEEEbiography}

\vspace{11pt}

\begin{IEEEbiography}[{\includegraphics[scale=0.12]{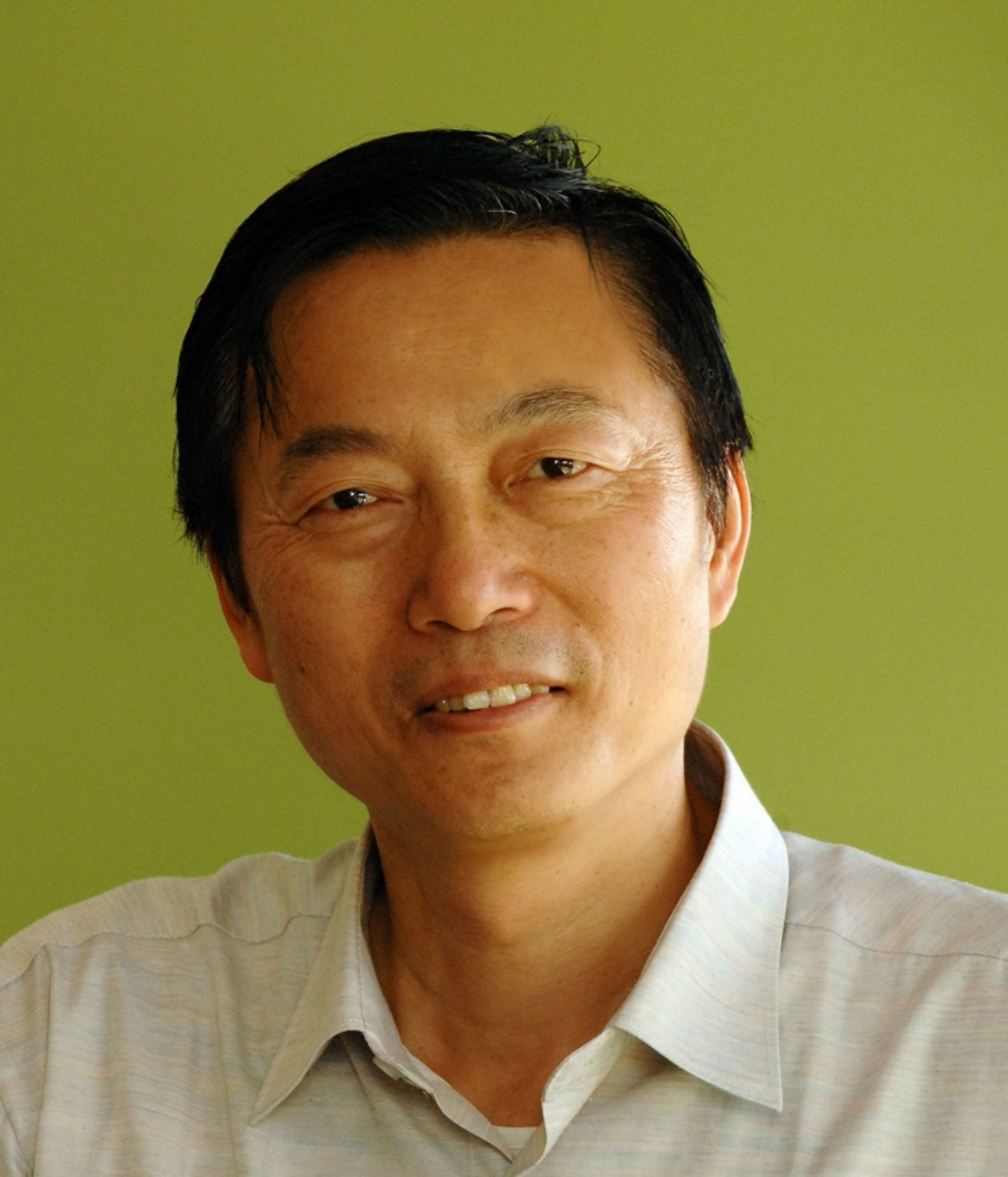}}]{Wanlei Zhou}
is currently the Vice Rector (Academic Affairs) and Dean of Institute of Data Science, City University of Macau, Macao SAR, China. He received the B.Eng and M.Eng degrees from Harbin Institute of Technology, Harbin, China in 1982 and 1984, respectively, and the PhD degree from The Australian National University, Canberra, Australia, in 1991, all in Computer Science and Engineering. He also received a DSc degree (a higher Doctorate degree) from Deakin University in 2002. Before joining City University of Macau, Professor Zhou held various positions including the Head of School of Computer Science in University of Technology Sydney, Australia, the Alfred Deakin Professor, Chair of Information Technology, Associate Dean, and Head of School of Information Technology in Deakin University, Australia. Professor Zhou also served as a lecturer in University of Electronic Science and Technology of China, a system programmer in HP at Massachusetts, USA; a lecturer in Monash University, Melbourne, Australia; and a lecturer in National University of Singapore, Singapore. His main research interests include security, privacy, and distributed computing. Professor Zhou has published more than 400 papers in refereed international journals and refereed international conferences proceedings, including many articles in IEEE transactions and journals. 
\end{IEEEbiography}

\vfill	
\end{document}